%% file: main.tex
\documentclass[10pt,twocolumn,letterpaper]{article}

\usepackage{cvpr}              

\input{preamble}

\definecolor{cvprblue}{rgb}{0.21,0.49,0.74}
\usepackage[pagebackref,breaklinks,colorlinks,citecolor=cvprblue]{hyperref}

\def\paperID{5997} 
\def\confName{CVPR}
\def\confYear{2025}

\title{Textured Gaussians for Enhanced 3D Scene Appearance Modeling}

\author{
Brian Chao \textsuperscript{1,2} \quad 
Hung-Yu Tseng \textsuperscript{2} \quad 
Lorenzo Porzi \textsuperscript{2} \quad 
Chen Gao \textsuperscript{2} \quad 
Tuotuo Li \textsuperscript{2} \quad 
Qinbo Li \textsuperscript{2} 
\\
Ayush Saraf \textsuperscript{2} \quad 
Jia-Bin Huang \textsuperscript{2,3} \quad 
Johannes Kopf \textsuperscript{2} \quad 
Gordon Wetzstein \textsuperscript{1} \quad 
Changil Kim \textsuperscript{2}
\\[2ex] 
{\textsuperscript{1} Stanford University} \quad
{\textsuperscript{2} Meta} \quad
{\textsuperscript{3} University of Maryland College Park}
\\[1ex]
{\small\bf \href{https://textured-gaussians.github.io}{\texttt{https://textured-gaussians.github.io}}}
}

\begin{document}
\twocolumn[{%
\renewcommand\twocolumn[1][]{#1}%
\maketitle
\begin{center}
    \centering
    \captionsetup{type=figure}
    \includegraphics[width=\textwidth]{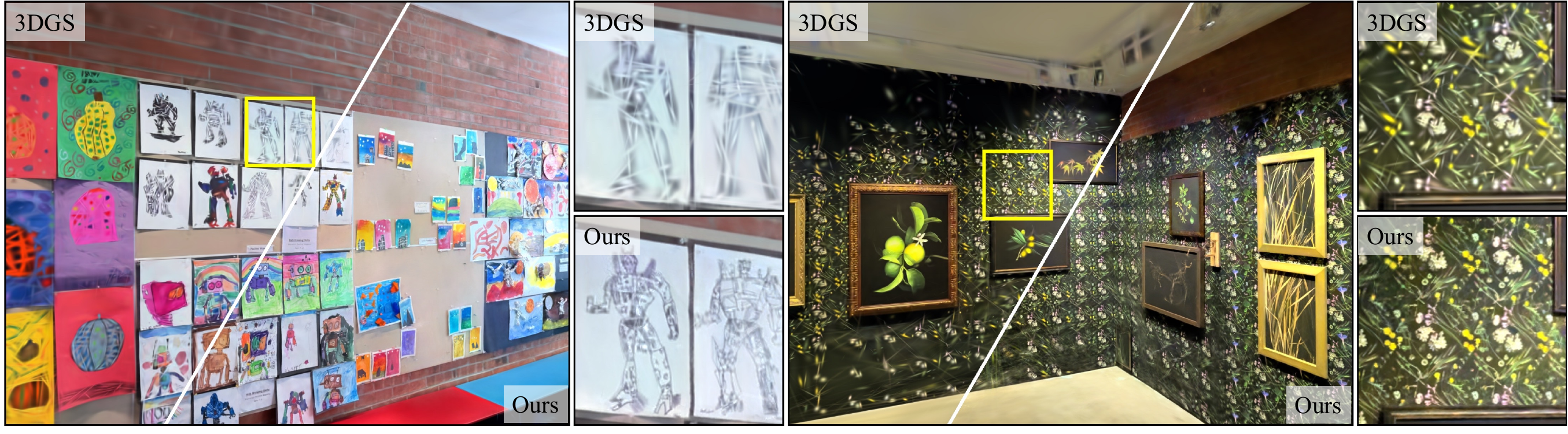}
    \captionof{figure}{\textbf{Textured Gaussians compared to 3D Gaussian Splatting (3DGS).}
    Our RGBA Textured Gaussians enhance 3D scene appearance modeling, leading to improved rendering quality while using the same number of Gaussians compared to 3DGS \cite{kerbl3Dgaussians}.
    Above, we show that Textured Gaussians faithfully reconstruct the fine details of scenes. 
    }
    \label{fig:teaser}
\end{center}%
}]

\input{sec/0_abstract}  
\vspace{-5mm}
\input{sec/1_intro}

\input{sec/2_related_work}
\input{sec/3_method}

\input{sec/4_experiments}
\input{sec/5_discussion}

{
    \small
    \bibliographystyle{ieeenat_fullname}
    \bibliography{main}
}

\input{sec/X_suppl}


\end{document}

%% file: preamble.tex
\usepackage[dvipsnames]{xcolor}

\usepackage{multirow}  
\usepackage{graphicx}
\usepackage{makecell}
\usepackage{bm} 
\usepackage{float} 
\usepackage{arydshln}  
\usepackage{hhline}
\usepackage{booktabs}
\usepackage{float}
\usepackage{placeins}  


%% file: sec/0_abstract.tex
\begin{abstract}

\noindent
3D Gaussian Splatting (3DGS) has emerged as the state-of-the-art 3D reconstruction technique, offering high-quality results with fast training and rendering. 
However, its expressivity is limited as pixels covered by the same Gaussian share identical colors aside from a Gaussian falloff scaling factor, and individual Gaussians can only represent simple ellipsoids geometrically. To overcome these limitations, we integrate texture and alpha mapping from traditional graphics with 3DGS. Our approach augments each Gaussian with alpha, RGB, or RGBA texture maps to model spatially varying color and opacity across each Gaussian's extent. This allows Gaussians to represent richer texture patterns and geometric structures beyond single-color ellipsoids. Notably, alpha-only texture maps significantly improve Gaussian expressivity, while further augmenting with RGB texture maps achieve maximum expressivity. We validate our method on a wide variety of standard benchmark datasets and our own custom captures at both the object and scene levels, and demonstrate image quality improvements over existing methods while using a similar or lower number of Gaussians.

\end{abstract}

%% file: sec/1_intro.tex
\section{Introduction}
\label{sec:intro}
Neural rendering \cite{tewari2022advances} achieves unprecedented novel-view synthesis and 3D reconstruction quality, offering solutions to a myriad of applications ranging from surface reconstruction \cite{li2023neuralangelo, wang2021neus}, SLAM \cite{rosinol2023nerfslam, nice2022CVPR}, material estimation and relighting \cite{nerfactor2021, nerv2021}, to virtual teleportation and human avatar animation \cite{weng2022humannerf, su2021anerf}. Recently, 3D Gaussian Splatting~(3DGS)~\cite{kerbl3Dgaussians} emerged as a state-of-the-art novel-view synthesis technique, with desirable properties such as high image quality results, fast training and rendering time, and an explicit primitives-based representation. 
Subsequent works in Gaussian Splatting have focused on topics such as improving surface reconstruction quality \cite{Huang2DGS2024, guedon2023sugar}, scene editing \cite{guedon2024frosting, chen2024gaussianeditor}, dynamic scene reconstruction \cite{ yang2023deformable3dgs,shih2024modeling}, and 3D generation \cite{tang2023dreamgaussian, ren2023dreamgaussian4d}.
Despite the success, 3DGS sometimes fails to model detailed appearances, as demonstrated in Figure~\ref{fig:teaser}.


In 3DGS, each Gaussian can only represent a single color and the shape of an ellipsoid given a camera viewpoint. 
This dramatically limits the set of appearances and shapes a single Gaussian can represent. 
To address the issue, \citet{huang2024spatial} modified the viewing direction calculation in 3DGS such that pixels covered by the same Gaussian exhibit a smooth color and opacity variation. 
However, this color variation across a single Gaussian is minuscule due to the spherical harmonics representation of colors, preventing them from representing complex textures. ~\citet{xu2024texturegs} proposed only using Gaussians to represent scene geometry and leveraged a learned UV mapping module and a global 2D RGB texture map to project textures onto Gaussian surfaces. 
However, the unit-sphere representation of textures significantly constrains the capacity of their model, and thus, their method fails to reconstruct objects with complex geometry or large-scale scenes.

\begin{figure}[!t]
    \centering
    \includegraphics[width=\linewidth]{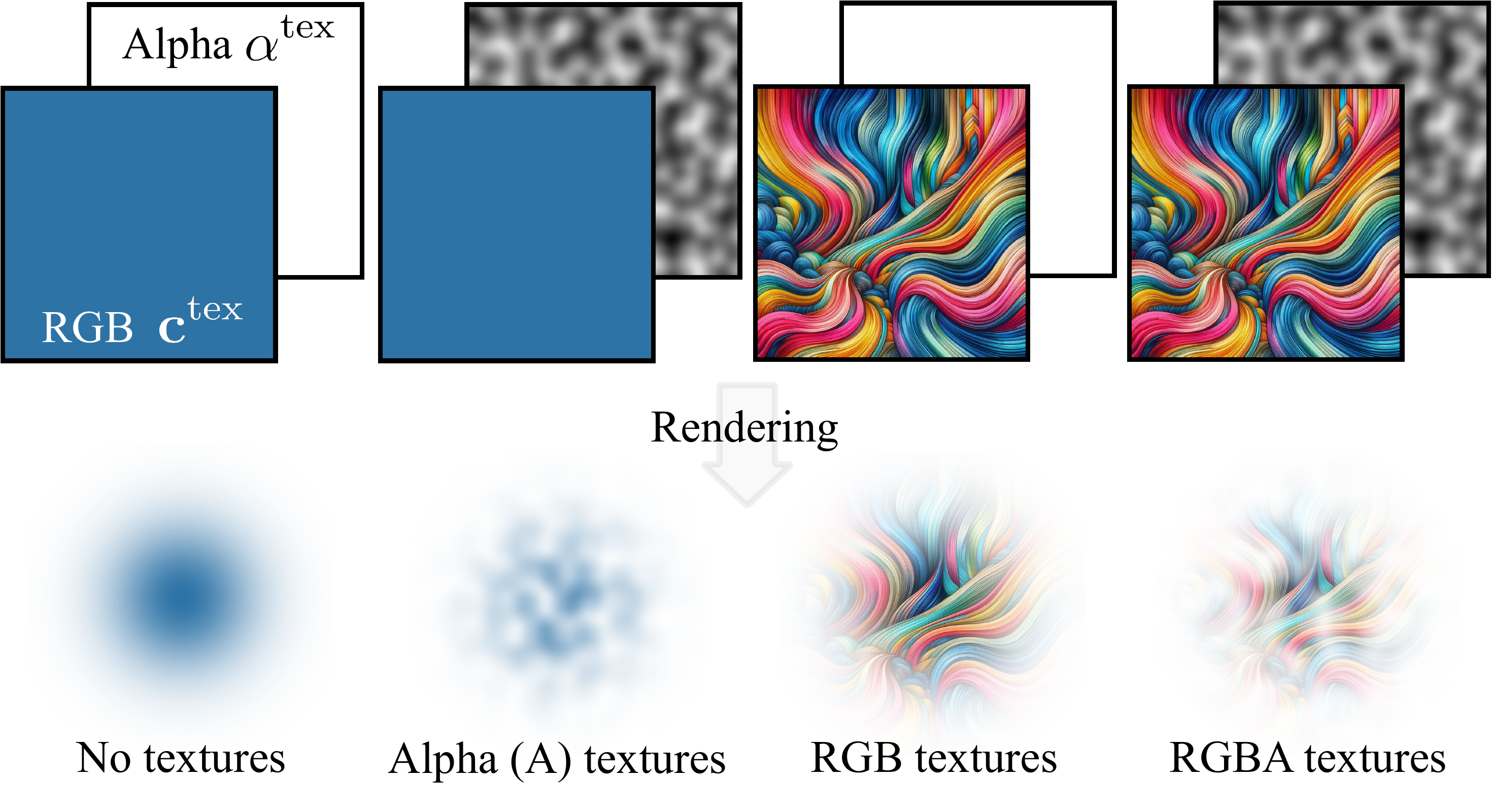}
    \caption{\textbf{Variants of our Textured Gaussians model.} 
    Textured Gaussians encapsulate four kinds of color and opacity spatial variations. 
    The top row of the figure shows the texture map associated with each Gaussian, and the bottom row shows the rendered Textured Gaussians. 
    The constant-color and constant-alpha model (no textures) corresponds to the original 3DGS formation, which can only represent a single color up to a Gaussian falloff factor within the Gaussian extent. 
    Textured Gaussians can already model spatially varying colors using only alpha textures since each pixel can be alpha-composited differently. 
    The model achieves maximum expressivity when leveraging the full RGBA texture map, where each Gaussian is capable of representing complex shapes and high frequency textures.}
    \label{fig:rgba_textures}
\end{figure}

Our method builds upon 3DGS and draws inspiration from mesh-based 3D representations that model appearance using texture mapping. 
As shown in Figure~\ref{fig:rgba_textures}, we augment each Gaussian with its alpha, RGB, or RGBA texture map so that each Gaussian is capable of representing a rich set of textures and shapes.
We call this representation~\emph{Textured Gaussians}. 
With RGB textures, individual Gaussians can represent higher-frequency color variations. 
The alpha maps allow Gaussians to represent a wider variety of shapes, instead of just ellipsoids. 
To achieve this, we build custom CUDA kernels that perform ray-Gaussian intersection and texture mapping and integrate them with 3DGS. 
Consequently, our method not only inherits all desirable properties of 3DGS, such as fast training and rendering time and explicit 3D representation, but also substantially improves rendered image quality, especially with a small number of Gaussians. 

In summary, our contributions include:
\begin{itemize}
    \item Introduction of a generalized appearance model for 3D Gaussians that handles spatially varying color and opacity by augmenting 3D Gaussians with alpha, RGB, or RGBA texture maps. 
    \item Validation that Textured Gaussians improves 3DGS on a wide variety of \textit{both} object-level and scene-level datasets.
    \item Demonstration that our method greatly outperforms 3DGS when the number of Gaussians is small and achieves better performance than 3DGS with alpha-only textures given the same model size.
\end{itemize}

%% file: sec/2_related_work.tex
\section{Related Work}

\noindent \textbf{Novel-View Synthesis (NVS) and Neural Rendering.} 
NVS tackles the problem of generating accurate renderings of 3D scenes from unseen viewpoints given a set of training images and camera poses. 
NVS traces back to the Structure from Motion (SfM) works \cite{snavely2026phototourism, Hartley2004sfm, brown2005unsupervised} and Multiview Stereo (MVS) \cite{seitz2006comparison, goesele2007multi, huang2018deepmvs}, which served as the basis of several groundbreaking NVS methods \cite{DeepBlending2018, eisemann2008floating, chaurasia2013depth, KPLD21point}. 
However, these approaches generally require storing hundreds and thousands of input images for reprojection and blending, which leads to massive memory requirements and sometimes fails to reconstruct undersampled regions. 

Recent advances in neural rendering \cite{tewari2022advances} have made great strides in improving the quality of 3D reconstruction and novel-view rendering. 
Neural rendering algorithms can be roughly classified by the underlying 3D representation, ranging from point clouds~\cite{KPLD21point, aliev2019point, xu2022point}, voxels~\cite{yu_and_fridovichkeil2021plenoxels, hu2023multiscale, yu2021plenoctrees}, meshes~\cite{neumesh, kato2018meshrenderer, Lombardi2018deepappearance, hu2021worldsheet, guo2023vmesh}, to implicit representations using MLPs~\cite{mildenhall2020nerf, mueller2022instant, sitzmann2019srns,attal2022learning}. 
Recently, 3D Gaussian Splatting \cite{kerbl3Dgaussians} (3DGS) has emerged as the state-of-the-art NVS technique. 
3DGS achieves high image quality for NVS, is fast to optimize and render, and leverages an explicit Gaussian primitive representation, which makes the method suitable for a wide variety of tasks such as surface extraction \cite{guedon2023sugar, Huang2DGS2024}, human avatar manipulation \cite{qian20233dgsavatar, shao2024splattingavatar}, ambient dynamic modeling~\cite{shih2024modeling}, and object and scene editing \cite{guedon2024frosting, gaussian_grouping}. 

\begin{figure*}[!t]
    \centering
\includegraphics[width=\textwidth]{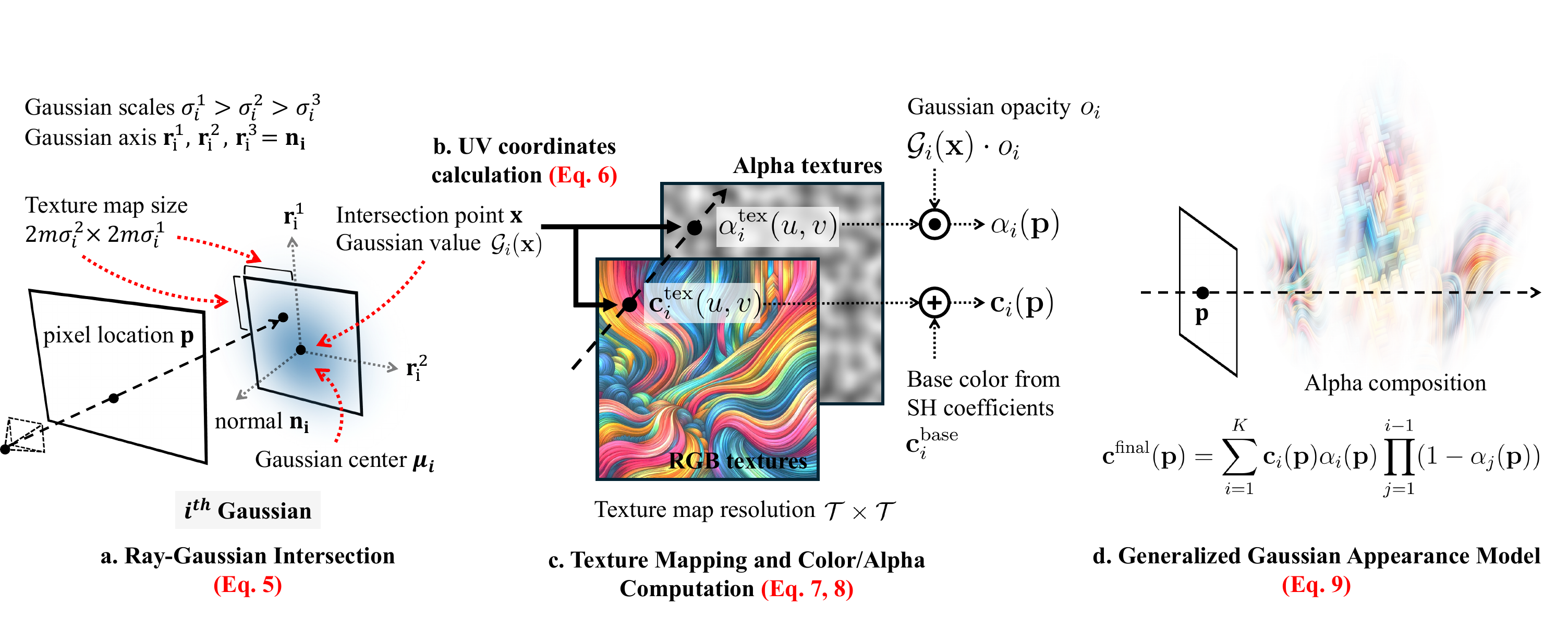}
    \caption{\textbf{Textured Gaussians model pipeline.} 
    Our method consists of three major components: ray-Gaussian intersection, RGBA texture mapping, and a generalized Gaussian appearance model. 
    To render the color of a pixel $\mathbf{p}$, we first trace a ray from the camera center $\mathbf{o}$ to the pixel to intersect with 3D Gaussians in the scene. 
    Then, we query texture and alpha values, $\mathbf{c}_\text{tex}$ and $\alpha_\text{tex}$, from the per-Gaussian RGBA texture maps using the ray-Gaussian intersection point $\mathbf{x}$. 
    Finally, given the retrieved spatially varying color and alpha values, we alpha-composite the color and alpha values of Gaussians that are hit by the pixel ray using the generalized Gaussian appearance model. }
    \label{fig:model_pipeline}
\end{figure*}

Our method augments 3DGS with alpha and texture mapping commonly used in mesh-based 3D representations. 
During optimization and rendering, we intersect outgoing rays from pixels with Gaussians in the scene. 
Intersection points are then used to query per-Gaussian texture map values via UV mapping and compute RGB color and alpha blending values. 
This enables each Gaussian to represent complex textures and shapes, significantly improving NVS quality.


\noindent \textbf{Appearance Modeling in 3D Gaussian Splatting.}  
In 3DGS, scene geometry is represented by the position, rotation, and scale of 3D Gaussians. By adjusting these properties, 3DGS can represent intricate geometric structures. 
On the other hand, appearance is modeled using per-Gaussian opacity values and spherical harmonics coefficients. However, pixels within a projected Gaussian are always shaded with the same color up to a Gaussian falloff factor, greatly limiting the expressivity of individual Gaussians. Therefore, recent works have explored ways that allow Gaussians to represent spatially varying features. \citet{huang2024spatial} defined per-Gaussian SH coefficients for color and opacity and used ray-Gaussian intersections to determine the viewing directions within the Gaussian extent. This enables smooth variation of colors and opacities across pixels covered by a single Gaussian but prevents the reconstruction of higher-frequency details. \citet{xu2024texturegs} disentangled appearance and geometry in 3DGS for texture editing, but their method is limited to object-centric scenes with simple geometry due to the unit-sphere parameterization of textures.

Instead of using a global texture map, we enhance each 3D Gaussian with a local texture map on top of the SH coefficients. 
This allows for a more flexible texture optimization since individual Gaussians are not tied to a shared global texture map. 
Our method can, therefore, reconstruct objects with complex structures and real-world scenes. 
Furthermore, each Gaussian can also represent various shapes with the alpha channel in the texture map.

\noindent \textbf{Memory Efficient Gaussian Splatting. } 
In order to achieve even faster rendering time and compact storage requirements for potential applications on edge devices, there has been a plethora of recent work focusing on optimizing memory-efficient Gaussian Splatting models \cite{bagdasarian20243dgszipsurvey3dgaussian}. 
These algorithms either prune Gaussians and perform quantization of Gaussian attributes \cite{kerbl2024reducing, lee2024compact, navaneet2023compact3d, morgenstern2023compact, fan2023lightgaussian, Niedermayr_2024_CVPR} or exploit the structural relationships between Gaussians \cite{chen2024hash, tao2024scaffoldgs}. 

Our work is orthogonal and complementary to these works since we focus on improving the appearance modeling of Gaussians by redistributing the model size budget to the per-Gaussian texture maps given any optimized 3DGS model. 
These model compression algorithms can be easily integrated with the 3DGS pretraining stage in our optimization pipeline to achieve further gains in compactness.

\noindent \textbf{Concurrent Work.} Different from several concurrent works that have explored the idea of augmenting Gaussians with texture maps \cite{rong2024gstex, song2024hdgstextured2dgaussian, xu2024SuperGaussians, svitov2025billboardsplattingbbsplatlearnable}, we 1) additionally utilize alpha textures and 2) provide an in-depth analysis of various texture map variants (alpha-only, RGB, RGBA) that explores the balance between reconstruction capability and model size.

%% file: sec/3_method.tex
\definecolor{light_red}{RGB}{247, 111, 111}
\definecolor{light_orange}{RGB}{247, 168, 111}
\definecolor{light_yellow}{RGB}{247, 208, 111}

\newcommand{\first}[1]{{\textbf{#1}}}
\newcommand{\second}[1]{{\underline{#1}}}
\newcommand{\third}[1]{#1}

\section{Method}

\subsection{3D Gaussian Splatting Model}
In 3DGS \cite{kerbl3Dgaussians}, 3D scenes are represented with 3D Gaussians, and images are rendered using differentiable volume splatting. Specifically, 3DGS explicitly defines 3D Gaussians by their 3D covariance matrix $\mathbf{\Sigma}_i \in \mathbb{R}^{3\times3} $ and center $\bm{\mu}_i \in \mathbb{R}^3$ (the index $i$ indicating the $i^\text{th}$ Gaussian), where the 3D Gaussian function value at point $\mathbf{x}\in \mathbb{R}^{3}$ is defined by:
\begin{equation}
    \mathcal{G}_i(\mathbf{x}) = \text{exp}(-\frac{1}{2}(\mathbf{x} - \bm{\mu}_i)\mathbf{\Sigma}_i^{-1}(\mathbf{x} - \bm{\mu}_i))
\end{equation}
where the covariance matrix $\mathbf{\Sigma} = \mathbf{R}\mathbf{S}\mathbf{S}^\top\mathbf{R}^\top$ is factorized into the rotation matrix $\mathbf{R} \in \mathbb{R}^{3\times3}$ and the scale matrix $\mathbf{S} \in \mathbb{R}^{3\times3}$. To render a 2D image from a 3D Gaussians representation, 3D Gaussians are transformed from world coordinates to camera coordinates via a world-to-camera transformation matrix $\mathbf{W} \in \mathbb{R}^{3\times3} $ and projected to the 2D image plane via a local affine transformation $\mathbf{J} \in \mathbb{R}^{3\times3} $. The transformed 3D covariance $\mathbf{\Sigma}'$ can be calculated as:
\begin{equation}
    \mathbf{\Sigma}' = \mathbf{J}\mathbf{W}\mathbf{\Sigma}\mathbf{W}^\top\mathbf{J}^\top
\end{equation}
The covariance $\Sigma_\text{2D}$ of the 2D Gaussian $\mathcal{G}^\text{2D}$ splatted on the image plane can be approximated as extracting the first two rows and columns of the transformed 3D covariance, $\Sigma_\text{2D} = [\Sigma']_{\{1,2\}, \{1,2\}} \in \mathbb{R}^{2\times2} $ using the matrix minor notation.

To render the color of a pixel $\mathbf{p}\in \mathbb{R}^3$, the colors associated with each Gaussian are alpha-composited from front to back following the conventional volume rendering equation:
\begin{equation}
    \mathbf{c}(\mathbf{p}) = \sum_{i=1}^K \mathbf{c}_i \alpha_i \prod_{j=1}^{i-1}(1 - \alpha_j) \label{eq:3dgs_color}
\end{equation}
where $i$ is the index of the Gaussians, $\mathbf{c}_i$ is the color of each Gaussian computed from the per-Gaussian spherical harmonic coefficients and viewing direction, and $\alpha_i$ is the alpha value computed from the opacity $o_i$ associated with each Gaussian and the 2D Gaussian value evaluated at pixel location $\mathbf{p}$:
\begin{equation}
    \alpha_i = \mathcal{G}^{\text{2D}}_i(\mathbf{p})\cdot o_i
\end{equation}

The attributes (center, rotation, scale, opacity, and spherical harmonic coefficients) of the Gaussians are optimized with gradient descent using photometric losses on the rendered 2D images.

\subsection{Textured Gaussians}
From the 3DGS appearance model defined in Eq. \ref{eq:3dgs_color}, we observe two properties of Gaussian Splatting:
\begin{enumerate}
    \item Pixels covered by the same Gaussian are shaded with the same color up to a Gaussian falloff scaling factor.
    \item The per-Gaussian opacity only allows Gaussians to represent ellipsoidal shapes.
\end{enumerate}
These two properties greatly restrict the expressivity of individual 3D Gaussian primitives. To allow each Gaussian primitive to represent complex appearances and shapes, we assign a fixed-resolution 2D texture map of size $\mathcal{T} \times \mathcal{T} \times \mathcal{K}, \mathcal{T} \in \mathbb{N}, \mathcal{K} \in \{1, 3, 4\}$ to each Gaussian and transform each Gaussian into a Textured Gaussian. As shown in Figure~\ref{fig:rgba_textures}, $\mathcal{K} = 1, 3, 4$ correspond to alpha, RGB, and RGBA texture maps, respectively. 
The shape of a Textured Gaussian is thus determined by the spatially varying opacity defined by the product of the per-Gaussian opacity and the alpha channel of the texture map. 
The appearance of a Textured Gaussian is represented by 1) the RGB channels of the \emph{spatially varying} textured map and 2) a set of \emph{spatially constant} spherical harmonic coefficients (SH).
Intuitively, the spatially constant SH coefficients represent the low-frequency textures plus view-dependent colors like specular effects,
while the spatially varying texture map represents the higher-frequency spatial variations of the texture. The texture is then mapped to the plane $\mathcal{P}$ defined by the two major axes of the $i^\text{th}$ 3D Gaussian that is centered at $\bm{\mu}_i \in \mathbb{R}^3 $, as shown in Figure \ref{fig:model_pipeline}. The normal $\mathbf{n}_i \in \mathbb{R}^3$ of this plane thus corresponds to the axis with the smallest scale. 

For each pixel $\mathbf{p} \in \mathbb{R}^3$ that is currently being rendered, we cast a ray from the camera origin $\mathbf{o} \in \mathbb{R}^3$ to the center of the pixel $\mathbf{p}$ and intersect it with the plane $\mathcal{P}$. The intersection point $\mathbf{x} \in \mathbb{R}^3$ can be calculated as:
\begin{equation}
    \mathbf{x} = \mathbf{o} + \frac{(\bm{\mu}_i - \mathbf{o}) \cdot \mathbf{n}_i}{(\mathbf{p} - \mathbf{o}) \cdot \mathbf{n}_i} \cdot \frac{\mathbf{p} - \mathbf{o}}{\|\mathbf{p} - \mathbf{o}\|}
\end{equation}
Given the intersection point $\mathbf{x}$ with $\mathcal{P}$, we perform UV mapping and bilinear interpolation on the texture map associated with the $i^\text{th}$ Gaussian to query the texture color and alpha value at the UV coordinates $u, v \in \mathbb{R}$, denoted by $\mathbf{c}_i^\text{tex}(u, v) \in \mathbb{R}^3$ and $\alpha_i^\text{tex}(u, v) \in \mathbb{R}$. Specifically, the UV coordinates $(u, v)$ of the intersection point on the texture map can be calculated as:

\begin{equation}
\begin{split}
    u = \frac{m\cdot \sigma_i^1 + (\mathbf{x} - \bm{\mu}_i)\cdot \mathbf{r}_i^1}{2\cdot m \cdot \sigma_i^1} \cdot (\mathcal{T} - 1) \\
    v = \frac{m\cdot \sigma_i^2 + (\mathbf{x} - \bm{\mu}_i)\cdot \mathbf{r}_i^2}{2\cdot m \cdot \sigma_i^2} \cdot (\mathcal{T} - 1)
\end{split}
\end{equation}
where $\sigma_i^1, \sigma_i^2 \in \mathbb{R}$ are the scales of the two major axis of the $i^\text{th}$ Gaussian, $\mathbf{r}_i^1, \mathbf{r}_i^2 \in \mathbb{R}^3$ are the normalized directions of the two major axes of the $i^\text{th}$ Gaussian, and $m \in \mathbb{R}$ is a scalar multiplier that determines the extent of the texture map with respect to each Gaussian.

{\renewcommand{\arraystretch}{1.2}
\begin{table*}[!t]
  \centering
  \footnotesize
  \begin{tabular*}{\textwidth}{@{\extracolsep{\fill}}lccccc@{}}
    \toprule
    \makecell{Method} & \makecell{Blender \cite{mildenhall2020nerf}} & \makecell{Mip-NeRF~360 \cite{mildenhall2020nerf}}  & \makecell{DTU \cite{jensen2014DTU}} & \makecell{Tanks and Temples \cite{Knapitsch2017tanks}} & \makecell{Deep Blending \cite{DeepBlending2018}} \\
    \midrule
    Mip-NeRF~360 \cite{barron2022mipnerf360} & 30.34 / 0.9510 / 0.0600 & \first{27.69} / \third{0.7920} / \third{0.2370} & \first{34.28} / 0.9641 / 0.0698 & \third{22.22} / \third{0.7590} / \third{0.2570} & \first{29.40} / \first{0.9010} / \first{0.2450}
 \\
    Instant-NGP \cite{mueller2022instant} & \third{32.20} / \third{0.9590} / \third{0.0550} & 25.30 / 0.6710 / 0.3710 & 23.58 / 0.7653 / 0.2603  & 21.72 / 0.7230 / 0.3300 & 23.62 / 0.7970 / 0.4230 \\
    \citet{xu2024texturegs} & 28.97 / 0.9380 / 0.0550 & -------- / -------- / -------- & 30.03 / -------- / 0.1440 & -------- / -------- / -------- & -------- / -------- / -------- \\
    3DGS* & \second{33.08} / \second{0.9671} / \second{0.0440} & \third{27.26} / \first{0.8318} / \second{0.1871} & \third{33.54} / \second{0.9697} / \first{0.0551} & \second{24.18} / \second{0.8541} / \second{0.1754} & \third{28.04} / \second{0.8940} / \third{0.2707}
 \\
    Ours & \first{33.24} / \first{0.9674} / \first{0.0428} & \second{27.35} / \second{0.8274} / \first{0.1858} & \second{33.61} / \first{0.9699} / \second{0.0556} & \first{24.26} / \first{0.8542} / \first{0.1684} & \second{28.33} / \third{0.8908} / \second{0.2699}
 \\
    \midrule
    3DGS* ($10\%$) & 31.47 / 0.9590 / 0.0594 & 25.77 / 0.7796 / 0.2860 & 32.71 / 0.9627 / 0.0811 & 22.82 / 0.8020 / 0.2728 & 27.64 / 0.8853 / 0.3101
 \\
    Ours ($10\%$) & \textbf{32.14} / \textbf{0.9629} / \textbf{0.0489} & \textbf{26.32} / \textbf{0.7976} / \textbf{0.2323} & \textbf{32.74} / \textbf{0.9661} / \textbf{0.0582} & \textbf{23.41} / \textbf{0.8259} / \textbf{0.2122} & \textbf{27.98} / \textbf{0.8898} / \textbf{0.2804}

 \\
    \hdashline
     3DGS* ($1\%$) & 26.89 / 0.9160 / 0.1165 & 22.37 / 0.6293 / 0.4774 & 30.88 / 0.9320 / 0.1581 & 19.90 / 0.6736 / 0.4406 & 23.97 / 0.8167 / 0.4337
 \\
    Ours ($1\%$) & \textbf{28.11} / \textbf{0.9343} / \textbf{0.0849} & \textbf{23.73} / \textbf{0.7064} / \textbf{0.3365} & \textbf{32.43} / \textbf{0.9627} / \textbf{0.0694} & \textbf{21.10} / \textbf{0.7399} / \textbf{0.3104} & \textbf{24.83} / \textbf{0.8454} / \textbf{0.3552}

 \\
    \bottomrule
  \end{tabular*}
  \caption{\textbf{Quantitative comparisons of different novel view synthesis methods.} We boldface the best-performing model and underline the second best in terms of different metrics (PSNR $\uparrow$ / SSIM $\uparrow$ / LPIPS $\downarrow$). Our Textured Gaussians model achieves better performance than 3DGS* across most metrics. With fewer Gaussians (1\% and 10\% of the default optimized number of Gaussians), our method significantly outperforms 3DGS*, achieving strictly better performance across all metrics as indicated by the bold text. }
  \label{tab:same_num_gs}
\end{table*}
}

Combined with the color computed from the SH coefficients, which we call $\mathbf{c}_i^\text{base}$ (the same as the color component in the original 3DGS appearance model in Eq.~\ref{eq:3dgs_color}), the final color contribution of the $i$-th Gaussian to pixel $\mathbf{p}$ is defined by:
\begin{equation}
    \mathbf{c}_i(\mathbf{p}) = \mathbf{c}_i^\text{base} + \mathbf{c}_i^\text{tex}(u, v)
    \label{eq:texture_map}
\end{equation}
and the alpha value of the $i$-th Gaussian at pixel $\mathbf{p}$ is defined by:
\begin{equation}
    \alpha_i(\mathbf{p}) = \alpha_i^\text{tex}(u, v) \cdot \mathcal{G}_i(\mathbf{x})\cdot o_i
    \label{eq:alpha_map}
\end{equation}
Finally, to render the color of a pixel $\mathbf{p}$, we modify the 3DGS appearance model in Eq.~\ref{eq:3dgs_color} to incorporate the spatially varying texture and opacity:
\begin{equation}
    \mathbf{c}^\text{final}(\mathbf{p}) = \sum_{i=1}^K \mathbf{c}_i(\mathbf{p}) \alpha_i(\mathbf{p}) \prod_{j=1}^{i-1}(1 - \alpha_j(\mathbf{p})) \label{eq:generalized_appearance_model}
\end{equation}
Eq.~\ref{eq:generalized_appearance_model} is a generalized formulation of 3D Gaussian appearance and encapsulates different variants of Textured Gaussians. For example, $\mathbf{c}_i^\text{tex} = 0$ and $\alpha_i^\text{tex} = 1$ correspond to the original 3DGS model.

\begin{figure}[!t]
  \centering
  \includegraphics[width=\linewidth]{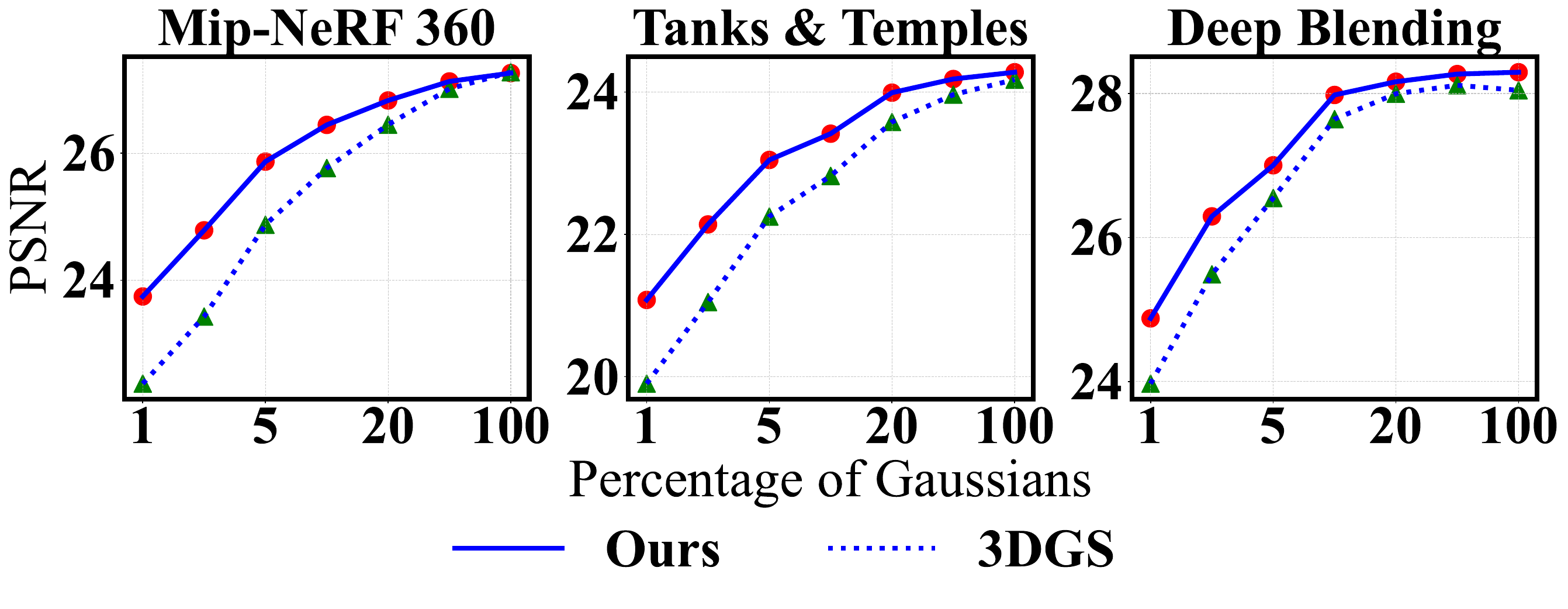}
  \caption{\textbf{NVS performance using varying numbers of Gaussians.} When using the same number of Gaussians as 3DGS$^*$, our Textured Gaussians achieve better novel view synthesis results on all five benchmark datasets in PSNR. Here, we show the quantitative performance of 3DGS* and Textured Gaussians models optimized with varying number of Gaussians on the three scene-level datasets.}
  \label{fig:vary_num_gs_trend}
\end{figure}

\subsection{Optimization of Textured Gaussians}
Following 3DGS \cite{kerbl3Dgaussians}, we optimize our model to minimize the weighted photometric loss:
\begin{equation}
    \mathcal{L} = \lambda \mathcal{L}_1 + (1 - \lambda) \mathcal{L}_\text{SSIM}
\end{equation}
where $\lambda = 0.8$. 

Our optimization procedure consists of two stages. We first optimize a 3DGS model for 30000 iterations. The learning rates and adaptive density control (ADC) parameters are reported in the Appendix. In the second stage, we initialize all attributes of the Gaussians with the optimized vanilla 3DGS model, and jointly optimize them with the per-Gaussian 2D texture maps for another 30000 iterations. We disable the ADC control in the second stage to control the number of Gaussians for fair baseline comparisons. This two-stage optimization process greatly speeds up convergence and improves image quality, since jointly optimizing all parameters is a highly ill-posed problem.

We implement custom CUDA kernels to perform fast ray-Gaussian intersection, UV mapping, and color composition. All experiments are conducted on clusters of Nvidia H100 GPUs. Please refer to the Appendix for more details on implementation and optimization.

%% file: sec/4_experiments.tex
\section{Results and Analysis}
We show selected qualitative and quantitative results in this section and refer the readers to the Appendix and the \href{https://textured-gaussians.github.io/}{project website} for an extensive set of results and video renderings.

\begin{figure*}[!t]
    \centering
\includegraphics[width=\textwidth]{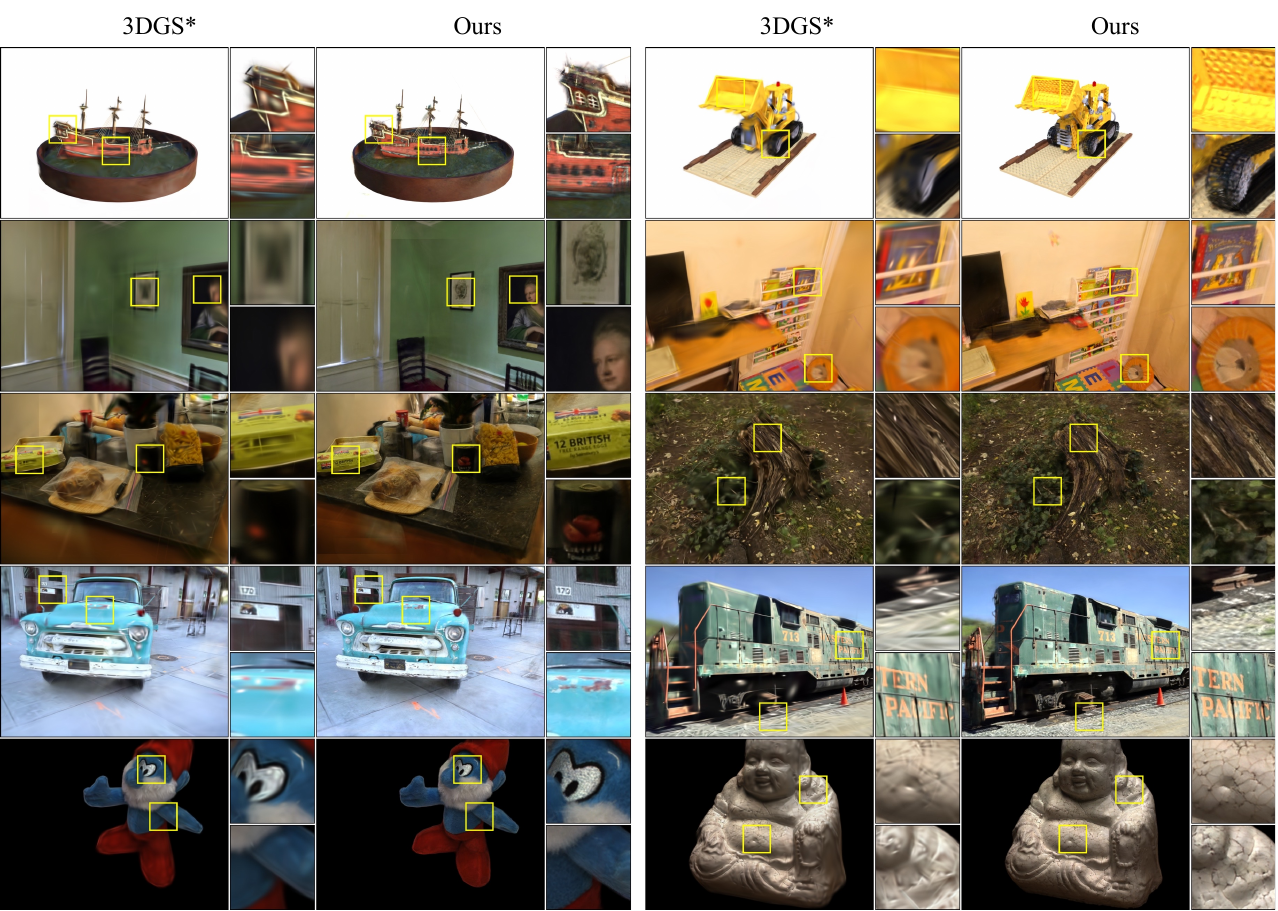}
    \caption{\textbf{Qualitative NVS results of benchmark datasets.} Novel-view rendering results for both object-level and scene-level standard benchmark datasets. Given the same small number of Gaussians (on average 2.5k and 39k Gaussians for object-level and scene-level datasets, respectively), 3DGS$^*$ fails to reconstruct high-frequency textures and complex shapes while our RGBA Textured Gaussians model succeeds. Please refer to the Appendix for results of our models optimized using varying number of Gaussians. }
    \label{fig:main_qualitative}
\end{figure*}

\noindent \textbf{Datasets and Evaluation Protocols. }
We evaluate the novel-view synthesis performance of our method on the 8 synthetic scenes from the Blender dataset \cite{mildenhall2020nerf}, all 9 scenes from the Mip-NeRF~360 dataset \cite{barron2022mipnerf360}, 5 scenes from the DTU dataset \cite{jensen2014DTU}, 2 scenes from the Tanks and Temples dataset \cite{Knapitsch2017tanks}, and 2 scenes from the Deep Blending dataset \cite{DeepBlending2018}. Please refer to the Appendix for a detailed description of the dataset preparation process.
 
For comparisons with Mip-NeRF~360 \cite{barron2022mipnerf360} and Instant-NGP \cite{mueller2022instant}, we refer to \citet{kerbl3Dgaussians} for results in scene-level datasets \cite{barron2022mipnerf360, Knapitsch2017tanks, DeepBlending2018}, and refer to NeRFBaselines \cite{kulhanek2024nerfbaselines} for results on the Blender dataset \cite{mildenhall2020nerf}. Results on the DTU dataset \cite{jensen2014DTU} are from our own runs of the authors' released code. Please refer to the Appendix for additional experimental details.

\citet{xu2024texturegs} pointed out in Section 5.4 of their paper that their algorithm cannot handle scene-level datasets and objects with complex structures (i.e. Blender dataset), thus they only reported results on the DTU dataset \cite{jensen2014DTU}. We report those numbers and additionally show results on the Blender dataset from GStex \cite{rong2024gstex} authors' own runs of the algorithm described in \citet{xu2024texturegs}.

In addition to standard benchmark test scenes, we also capture our own set of scenes that contain artworks with highly detailed textures to help demonstrate the effectiveness of our method.

\noindent \textbf{3DGS Baseline. } We build our method on top of our own implementation of a slightly modified 3DGS that closely follows the algorithm described in Gaussian Opacity Fields \cite{Yu2024GOF}. 
Specifically, we use a ray-based formulation that allows for the exact evaluation of 3D Gaussian values without approximating splatted 3D Gaussians as 2D Gaussians. In addition, we used a revised densification strategy~\cite{Yu2024GOF,ye2024absgs}.
For fair comparisons with our method, we report the performance of our 3DGS implementation and label it as 3DGS$^*$ throughout the results section, and refer the readers to the Appendix for quantitative results reported in the original 3DGS paper \cite{kerbl3Dgaussians}.

\subsection{Quantitative Results}
In Table \ref{tab:same_num_gs}, we show quantitative comparisons in terms of PSNR/SSIM/LPIPS between our Textured Gaussians model using full RGBA textures and 3DGS$^*$ with the same number of Gaussians, and results of other baseline methods. Our method generally achieves better results than 3DGS$^*$ and all other methods, since Texture Gaussians are more expressive than vanilla 3D Gaussians due to the added texture maps. 

\begin{figure*}[!t]
    \centering
\includegraphics[width=\textwidth]{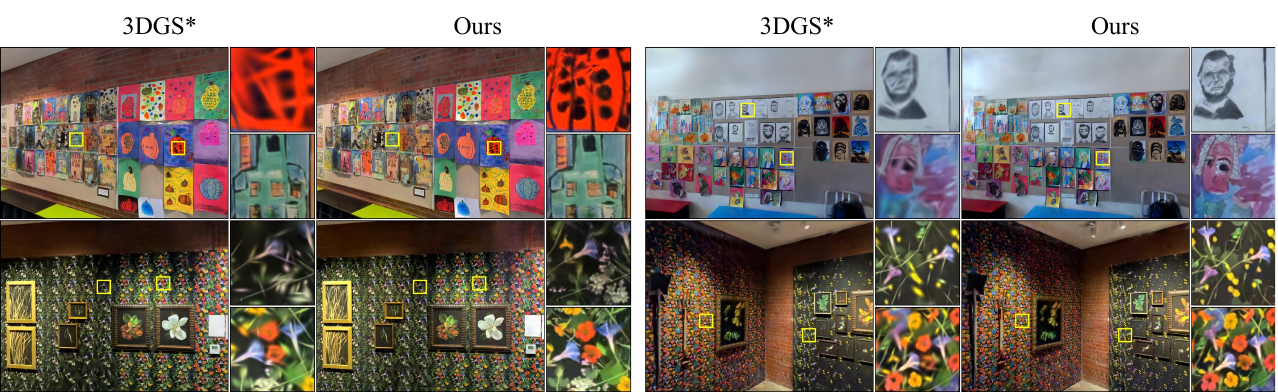}
    \caption{\textbf{Qualitative NVS results of custom datasets.} Our RGBA Textured Gaussians model achieves sharper reconstruction compared to 3DGS$^*$ when using the same small number of Gaussians (on average around 100k Gaussians). }
    \label{fig:custom_captures}
\end{figure*}

In the bottom half of Table \ref{tab:same_num_gs}, we also show comparisons between our method and 3DGS$^*$ when using a fraction of the default number of Gaussians, denoted by the percentage in parentheses. 
For our models with different numbers of Gaussians, we distribute a fixed amount of texels to all Gaussians (i.e., the same memory overhead for all Textured Gaussians models compared with 3DGS$^*$). Therefore, the texture map resolution of models with fewer Gaussians will be larger. We see that our method truly outshines 3DGS$^*$ with fewer Gaussians, achieving nearly 2 dB improvements over 3DGS$^*$ when using $1\%$ of the default number of Gaussians. Figure \ref{fig:vary_num_gs_trend} shows the trend of the quality of novel view synthesis as we vary the number of Gaussians. Our method consistently outperforms 3DGS$^*$ when using different numbers of Gaussians.

Table \ref{tab:same_model_size} compares the performance of 3DGS$^*$ and our Textured Gaussians model with the same model size using alpha-only textures and RGBA textures. Since our models use per-Gaussian texture maps, they require fewer Gaussians than 3DGS$^*$ to achieve the same model size. Our alpha-only textures model generally outperforms the 3DGS$^*$ and RGBA textures models, indicating that there is a sweet spot for distributing the model size budget between Gaussian parameters and texture map channels. This is further explored in Section~\ref{sec:exp_ablation}.


\begin{table}[!t]
  \centering
  \footnotesize
  \begin{tabular*}{\linewidth}{@{\extracolsep{\fill}}l@{\hskip 0.1cm}c@{\hskip 0.1cm}c@{\hskip 0.1cm}c@{\hskip 0.1cm}c@{\hskip 0.1cm}c@{\hskip 0.1cm}c@{}}
    \toprule
    Method & \multicolumn{2}{c}{3DGS$^*$} & \multicolumn{2}{c}{Alpha-only} & \multicolumn{2}{c}{RGBA} \\
    & PSNR & \#GS (M) & PSNR & \#GS (M)& PSNR & \#GS (M) \\
    \midrule
    Blender & \second{33.08} & 0.27 & \first{33.12} & 0.19 & 33.03 & 0.21 \\
    Mip-NeRF~360  & 27.26 & 4.4 & \first{27.37} & 3.1 & \second{27.26} & 3.5 \\
    DTU & \first{33.53} & 0.28 & \second{33.45} & 0.24 & 33.41 & 0.22 \\
    Tanks and Temples & 24.17 & 2.9 & \first{24.38} & 2.6 & \second{24.28} & 1.3 \\
    Deep Blending & 28.04 & 2.8 & \second{28.36} & 1.6 & \first{28.52} & 1.0 \\
    \bottomrule
  \end{tabular*}
  \caption{\textbf{Same model size comparisons.} We compare the PSNR performance between 3DGS$^*$ and our model using alpha-only and RGBA textures with the same model size. The average number of optimized Gaussians (in millions) from each dataset is shown in the \#GS (M) columns. We boldface and underline the best and second-best performing models, respectively. Alpha-only texture models generally achieve better performance than 3DGS$^*$ and RGBA texture models with the same model size.}
  \label{tab:same_model_size}
\end{table}

\subsection{Qualitative Results}
\noindent \textbf{Novel-View Synthesis. } We show novel view synthesis results of our model with RGBA textures and 3DGS$^*$ on both standard benchmark and custom-captured datasets in Figures \ref{fig:main_qualitative} and \ref{fig:custom_captures}. Our method reconstructs much sharper details of the scene compared to 3DGS$^*$ when using the same small number of Gaussians. Please refer to the Appendix for results of our models optimized using varying number of Gaussians.

\begin{figure*}[!t]
    \centering
\includegraphics[width=\textwidth]{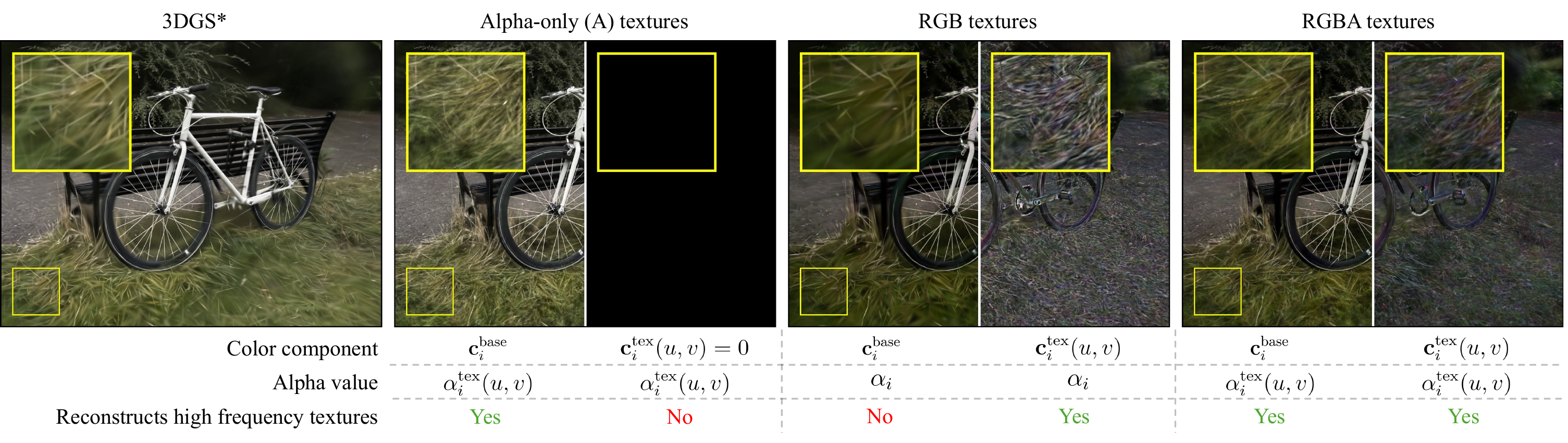}
    \caption{\textbf{Color component decomposition visualization.} Visualization of \textit{alpha-modulated and composited} $\mathbf{c}^\text{base}$ and $\mathbf{c}^\text{tex}$ color components in our Textured Gaussians model (Eq.~\ref{eq:texture_map}, \ref{eq:alpha_map}, \ref{eq:generalized_appearance_model}). Insets with yellow borders show zoom-in crops. Texture color brightness is adjusted for visualization. While 3DGS$^*$ fails to capture fine details with limited Gaussians, our method reconstructs sharp textures. With RGB textures, base color reconstructs low-frequency textures and texture maps reconstruct high-frequency elements. With alpha-only and RGBA textures, base color also reconstructs high-frequency textures via spatially varying alpha compositing.}
    \label{fig:color_contributions}
\end{figure*}

\noindent \textbf{Color Component Decomposition. }  We show the \textit{alpha-modulated and composited} results of the two color components, $\mathbf{c}^\text{tex}$ and $\mathbf{c}^\text{base}$, of the optimized alpha-only, RGB and RGBA textured Gaussians model with 1\% of the default number of Gaussians in Figure \ref{fig:color_contributions}. A 3DGS$^*$ model with the same number of Gaussians is shown for comparison. For RGB and RGBA models, $\mathbf{c}^\text{tex}$ reconstructs fine-grained textures that 3DGS$^*$ cannot. For alpha-only and RGBA models, $\mathbf{c}^\text{base}$ also reconstructs high-frequency details due to spatially varying alpha composition enabled by alpha textures.

\begin{figure}[!t]
    \centering
    \includegraphics[width=\linewidth]{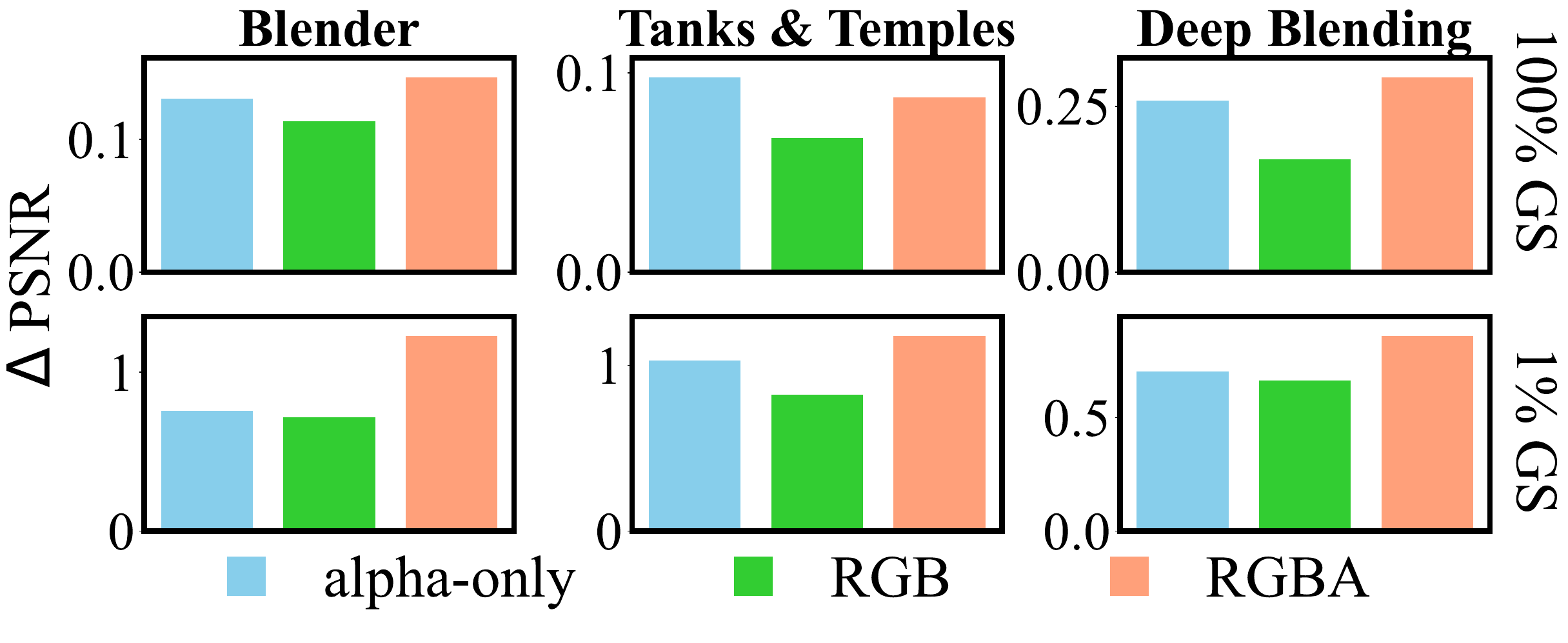}
    \caption{\textbf{Ablation study on texture map variants.} We show the PSNR performance improvements ($\Delta \text{PSNR}$) over 3DGS$^*$ of our Textured Gaussians model when using different texture map variants. Models with full RGBA textures achieve the best results. Interestingly, alpha-only textures outperform RGB textures and perform comparably to RGBA textures while using one-third and one-fourth of the model size, respectively. }
    \label{fig:texture_ablation}
\end{figure}

\subsection{Ablation Studies}
\label{sec:exp_ablation}

\noindent \textbf{Texture Map Variants. } 
We ablate variants of our model that use different texture maps, namely alpha-only, RGB, and RGBA texture maps, and the same number of Gaussians in Figure \ref{fig:texture_ablation}. Experiments are conducted on Blender, Tanks and Temples, and Deep Blending datasets with the default optimized number (top row) and 1\% of the default optimized number (bottom row) of Gaussians. We see that our model achieves the best performance with RGBA textures. Interestingly, using alpha-only textures already outperforms 3DGS$^*$, and our RGB textures model despite being one-third the size, striking a better balance between performance and model size. This is because alpha-textured Gaussians can represent complex shapes through spatially varying opacity and reconstruct high-frequency textures through spatially varying alpha composition. In contrast, RGB-textured Gaussians are still limited to representing ellipsoids.

\begin{figure}
    \centering
    \includegraphics[width=\linewidth]{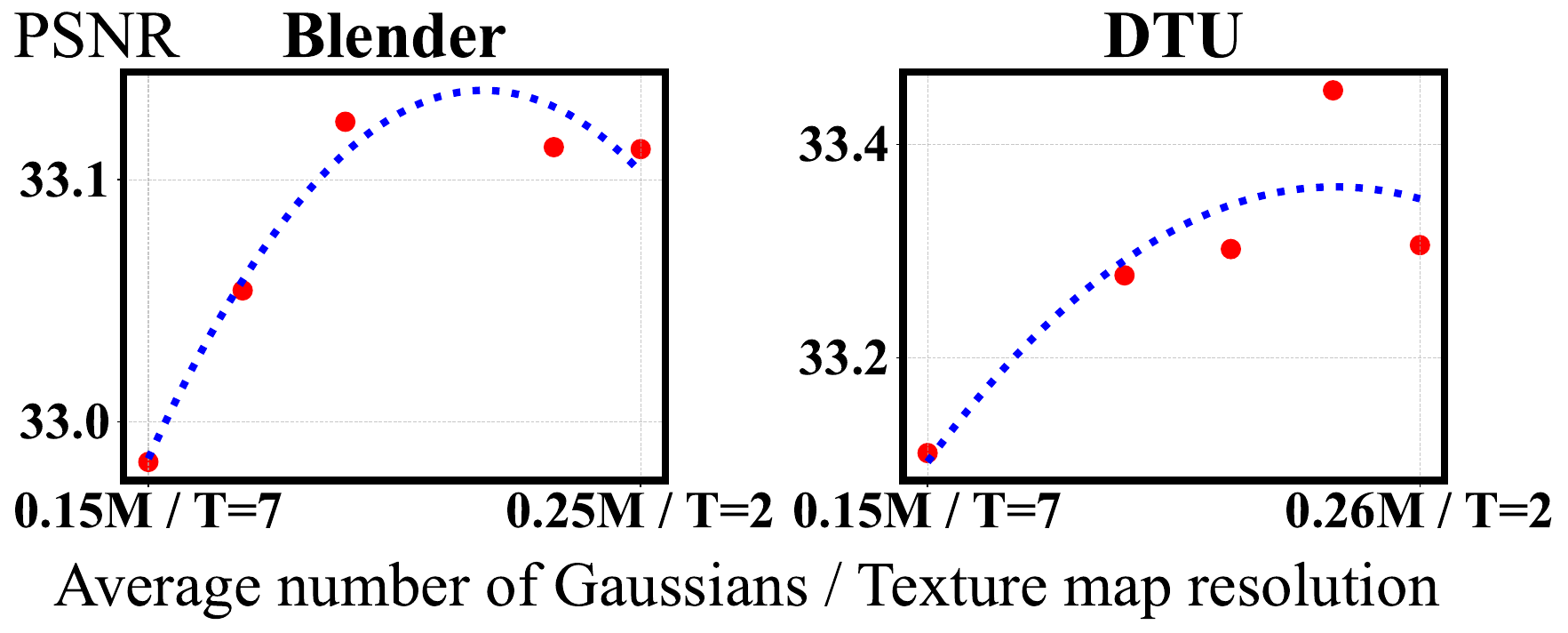}
    \caption{\textbf{Ablation study on texture map resolution and the number of Gaussians given the same model size.} We optimized various alpha-only Textured Gaussians models (red dots) with equivalent model sizes but different texture map resolutions and numbers of Gaussian. A fitted quadratic curve (blue dashed line) illustrates the performance trend. The optimal model variant balances these two parameters to achieve superior novel view synthesis results.}
    \label{fig:ablation_same_model_size}
\end{figure}

\noindent \textbf{Texture Map Resolution and the Number of Gaussians. } In Figure \ref{fig:ablation_same_model_size}, we optimize alpha-only Textured Gaussians models of the same model size with different texture map resolutions and number of Gaussians, and show the trend of novel view synthesis performance on the Blender and DTU datasets. We observe a sweet spot between texture map resolution and number of Gaussians that achieves the best image reconstruction quality. Maximizing texture map resolution corresponds to greatly reducing the number of Gaussians, which harms reconstruction of detailed geometric structures. Minimizing the texture map resolution, on the other hand, simply degenerates into 3DGS$^*$.

%% file: sec/5_discussion.tex
\section{Discussions}

\noindent \textbf{Limitations.} 
The use of 2D diffuse texture maps in our model assumes that all textures lie on a local surface and does not model spatially varying specular color. Thus, extending our texture representation to represent local 3D volume textures or even 5D radiance fields is crucial. Using factorized representations such as TensoRF \cite{Chen2022ECCVtensorf} or triplane \cite{Chan2021eg3d} to represent these high-dimensional textures could also be an interesting research problem. Finally, extending Textured Gaussians to support dynamic scene reconstruction \cite{yang2023deformable3dgs, shih2024modeling} through time-varying texture maps could be a natural next-step.

\noindent \textbf{Conclusions.} 
In this paper, we augment 3DGS with texture maps to allow individual Gaussians to model spatially varying colors and opacity. As such, each Gaussian can represent a much richer set of appearances and shapes. This greatly improves the expressivity of individual Gaussians, leading to better novel-view synthesis quality. Our method achieves better quality than 3DGS when using the same number of Gaussians, and achieves better or comparable quality when using the same model size. 

\section*{Acknowledgment}
We thank artist \href{https://www.kijalucas.com/}{Kija Lucas} for allowing us to use her art installation at the Palo Alto Art Center as one of our custom captured scenes (the \textit{flower gallery} scene). Brian Chao is supported by the Stanford Graduate Fellowship and the NSF GRFP.

%% file: sec/X_suppl.tex
\clearpage
\setcounter{page}{1}
\maketitlesupplementary
\setcounter{section}{0} 
\setcounter{figure}{0}
\setcounter{equation}{0}
\setcounter{table}{0}
\renewcommand{\thesection}{\Alph{section}}
\renewcommand{\thefigure}{A.\arabic{figure}}
\renewcommand{\theequation}{A.\arabic{equation}}
\renewcommand{\thetable}{A.\arabic{table}}

{\renewcommand{\arraystretch}{1.2}
\begin{table*}[!t]
  \centering
  \footnotesize
  \begin{tabular*}{\textwidth}{@{\extracolsep{\fill}}lccccc@{}}
    \toprule
    \makecell{Method} & \makecell{Blender \cite{mildenhall2020nerf}} & \makecell{Mip-NeRF 360 \cite{mildenhall2020nerf}}  & \makecell{DTU \cite{jensen2014DTU}} & \makecell{Tanks and Temples \cite{Knapitsch2017tanks}} & \makecell{Deep Blending \cite{DeepBlending2018}} \\
    \midrule
    \citet{kerbl3Dgaussians} & 33.32 / -------- / -------- & 27.21 / 0.815 / \underline{0.214} & -------- / -------- / -------- & 23.14 / 0.841 / \underline{0.183} & 29.41 / 0.903 / \underline{0.243} \\
    3DGS$^*$ & 33.09 / 0.967 / 0.044 & 27.28 / 0.832 / 0.187 & 33.54 / 0.970 / 0.055 & 24.18 / 0.854 / 0.175 & 28.04 / 0.894 / 0.271
\\
\bottomrule
  \end{tabular*}
  \caption{Quantitative performance comparison (PSNR $\uparrow$ / SSIM $\uparrow$ / LPIPS $\downarrow$) between our own 3DGS implementation (labelled as 3DGS* in the main paper) and the original 3DGS paper \cite{kerbl3Dgaussians}. The under-estimated LPIPS values from the original 3DGS paper are underlined. }
  \label{tab:original_3dgs}
\end{table*}
}

\section{Results from Original 3DGS Paper}
We show the quantitative results (PSNR / SSIM / LPIPS) of all 5 datasets reported in the original 3DGS paper \cite{kerbl3Dgaussians} and the performance of our own 3DGS implementation in Table \ref{tab:original_3dgs}. Since we used a modified version of 3DGS described in Gaussian Opacity Fields \cite{Yu2024GOF}, there are slight differences in the results. We compare our textured Gaussians model with our own modified 3DGS implementation (which we label as 3DGS$^*$ in the main paper) for a fair comparison, since our algorithm is built on top of that. However, the concept of Textured Gaussians could also be easily applied to the original 3DGS model. 

\textit{A note on LPIPS.} The LPIPS values reported in the original 3DGS paper \cite{kerbl3Dgaussians} are underestimated, leading to additional performance discrepancies compared to our method. This issue was pointed out in \cite{bulo2024revising} and confirmed in private correspondence with the authors of 3DGS \cite{kerbl3Dgaussians}. To avoid confusion, we underline the underestimated LPIPS values in Table \ref{tab:original_3dgs} and report the correct LPIPS values of our own implementation.

\section{Custom 3DGS Implementation Details}

Our custom 3DGS implementation closely follow the modified 3DGS algorithm described in Gaussian Opacity Fields \cite{Yu2024GOF}. In this section, we describe how each component in our own implementation of 3DGS differs from the original 3DGS paper \cite{kerbl3Dgaussians}.

\noindent \textbf{Gaussian Value Calculation. } In 3DGS, 3D scenes are represented with 3D Gaussians, and images are rendered using differentiable volume splatting. Specifically, 3DGS explicitly defines 3D Gaussians by their 3D covariance matrix $\mathbf{\Sigma}_i \in \mathbb{R}^{3\times3} $ and center $\bm{\mu}_i \in \mathbb{R}^3$ (the index $i$ indicating the $i^\text{th}$ Gaussian), where the 3D Gaussian function value at point $\mathbf{x}\in \mathbb{R}^{3}$ is defined by:
\begin{equation}
    \mathcal{G}_i(\mathbf{x}) = \text{exp}(-\frac{1}{2}(\mathbf{x} - \bm{\mu}_i)\mathbf{\Sigma}_i^{-1}(\mathbf{x} - \bm{\mu}_i))
\end{equation}
where the covariance matrix $\mathbf{\Sigma} = \mathbf{R}\mathbf{S}\mathbf{S}^\top\mathbf{R}^\top$ is factorized into the rotation matrix $\mathbf{R} \in \mathbb{R}^{3\times3}$ and the scale matrix $\mathbf{S} \in \mathbb{R}^{3\times3}$. When rendering the color of a pixel $\mathbf{p}\in \mathbb{R}^{3}$, 3D Gaussians are transformed from world coordinates to camera coordinates via a world-to-camera transformation matrix $\mathbf{W} \in \mathbb{R}^{3\times3} $ and projected to the 2D image plane via a local affine transformation $\mathbf{J} \in \mathbb{R}^{3\times3} $. The transformed 3D covariance $\mathbf{\Sigma}'$ can be calculated as:
\begin{equation}
    \mathbf{\Sigma}' = \mathbf{J}\mathbf{W}\mathbf{\Sigma}\mathbf{W}^\top\mathbf{J}^\top
\end{equation}

The covariance $\Sigma_\text{2D}$ of the 2D Gaussian $\mathcal{G}^\text{2D}$ splatted on the image plane can be \emph{approximated} as extracting the first two rows and columns of the transformed 3D covariance, $\Sigma_\text{2D} = [\Sigma']_{\{1,2\}, \{1,2\}} \in \mathbb{R}^{2\times2} $ using the matrix minor notation. The alpha value $\alpha_i$ used to perform alpha compisting at the pixel location can then be calculated by:
\begin{equation}
    \alpha_i = \mathcal{G}^{\text{2D}}_i(\mathbf{p})\cdot o_i
\end{equation}
where $o_i$ is the opacity of the $i^\text{th}$ 3D Gaussian.

Instead of \emph{approximating} 2D Gaussian values by splatting 3D Gaussians, we cast rays from the pixel currently being rendered and calculate the \emph{exact} 3D Gaussian value evaluated at the ray-Gaussian intersection point. This forumlation fits nicely with our Textured Gaussians algorithm since ray-Gaussian intersection calculation is also required from texture UV-mapping. The alpha value $\alpha_i$ used for alpha blending at pixel $\mathbf{p}$ is therefore defined by:
\begin{equation}
    \alpha_i = \mathcal{G}_i(\mathbf{x})\cdot o_i
\end{equation}
where $\mathbf{x}$ is the ray-Gaussian intersection point and $\mathcal{G}_i$ is the $i^\text{th}$ 3D Gaussian. 

\noindent \textbf{Adaptive Density Control (ADC). } The optimization of 3DGS starts from  a sparse structure-from-motion (SfM) point cloud and progressively densifies Gaussians either through cloning or splitting. In the original 3DGS, this densification process is guided by a score that is defined by the magnitude of the view/screen-space positional gradient $\frac{dL}{d\bm{\mu}'}$ of the Gaussian:
\begin{equation}
    \left\| \frac{dL}{d\bm{\mu}'} \right\|= \left\|\sum_j \frac{dL}{d\mathbf{p}_j} \cdot \frac{d\mathbf{p}_j}{d\bm{\mu}'}\right\|
\end{equation}
where $\mu'$ is the center of the projected Gaussian and $\mathbf{p}_j$ are the pixels the Gaussian contributed to. If this score is larger than a predfined threshold $\tau$, the Gaussian will be cloned or splitted.

As pointed out in Gaussian Opacity Fields \cite{Yu2024GOF}, this score is not effective in identifying overly blurred areas for Gaussian densification, and the authors proposed an alternative \emph{sum-of-magnitudes} score to replace the original magnitude-of-sums:
\begin{equation}
    \sum_j \left\|\frac{dL}{d\mathbf{p}_j} \cdot \frac{d\mathbf{p}_j}{d\bm{\mu}'}\right\|
\end{equation}

We use this revised densification score to perform adaptive density control throughout optimization. 

\noindent \textbf{Learning Rates and other Hyperparameters. } We use all default learning rates and hyperparameters in the original 3DGS implementation \cite{kerbl3Dgaussians}. Please refer to their released code for the specific values. For each Textured Gaussians experiment and ablation study, we use the same texture map resolution and set the texture map learning rate to be $0.001$ for all datasets and do not require per-dataset tuning for texture map parameters and learning rates.

\section{Additional Details on Implementation, Optimization, and Dataset Preparation} 

\noindent \textbf{Implementation Details.} We implement our algorithm using PyTorch and custom CUDA kernels that perform fast ray-Gaussian intersection calculation, UV mapping, and texture lookup. All experiments are run on clusters of Nvidia H100 GPUs. 

To avoid integer overflows during UV mapping, we only intersect Gaussians within $\pm 3\sigma$ along the two major axes and map the 2D texture map to the $\pm 3\sigma$ range ($m = 3$). The color of intersection points outside of this range is simply assigned as black, that is $c_i^\text{tex}(u, v) = 0$. We found that this implementation greatly improves numerical stability, especially when viewing Gaussians from grazing angles. This also has very little effect on the final performance since the opacity of the Gaussian outside of the $\pm 3\sigma$ range is negligible (less than 0.01). We also found that dynamically adjusting the intersection plane based on the two longest axes of the ellipsoid leads to empirically better performance.

\noindent \textbf{Optimization Details. } We set the learning rate of the texture maps to 0.001 and initialize the RGB colors of the texture map to a low value ($\frac{25}{255}$) since the optimized spherical harmonic coefficients of the first stage should have already learned to reconstruct the average color of the pixels within the Gaussian extent, and the texture map should only learn to reconstruct the residual color in the second stage of optimization. The alpha channel of the texture map is initialized to 1. 

Since our training procedure has two stages (3DGS pretraiing and Textured Gaussians optimization), the training time of our models are roughly two times the training time of 3DGS. Our custom CUDA kernels that perform ray-Gaussian intersection and texture lookup adds very little overhead to the original 3DGS rendering process, hence the inference time of our Textured Gaussians model is approximately the same as that of 3DGS.

\noindent \textbf{Dataset Preparation Details. } We downscale the DTU dataset to $800 \times 600$ image resolution following \citet{xu2024texturegs}, and conform to the dataset preparation and evaluation protocols described in \citet{kerbl3Dgaussians} for the remaining four datasets (Blender \cite{mildenhall2020nerf}, Mip-NeRF 360 \cite{barron2022mipnerf360}, Tanks and Temples \cite{Knapitsch2017tanks}, and Deep Blending \cite{DeepBlending2018}). Following
\citet{mildenhall2020nerf} and \citet{xu2024texturegs}, we use the alpha channel of the images to create black and white backgrounds for the DTU dataset and the Blender dataset, respectively.

\begin{figure}[!t]
    \centering
    \includegraphics[width=\linewidth]{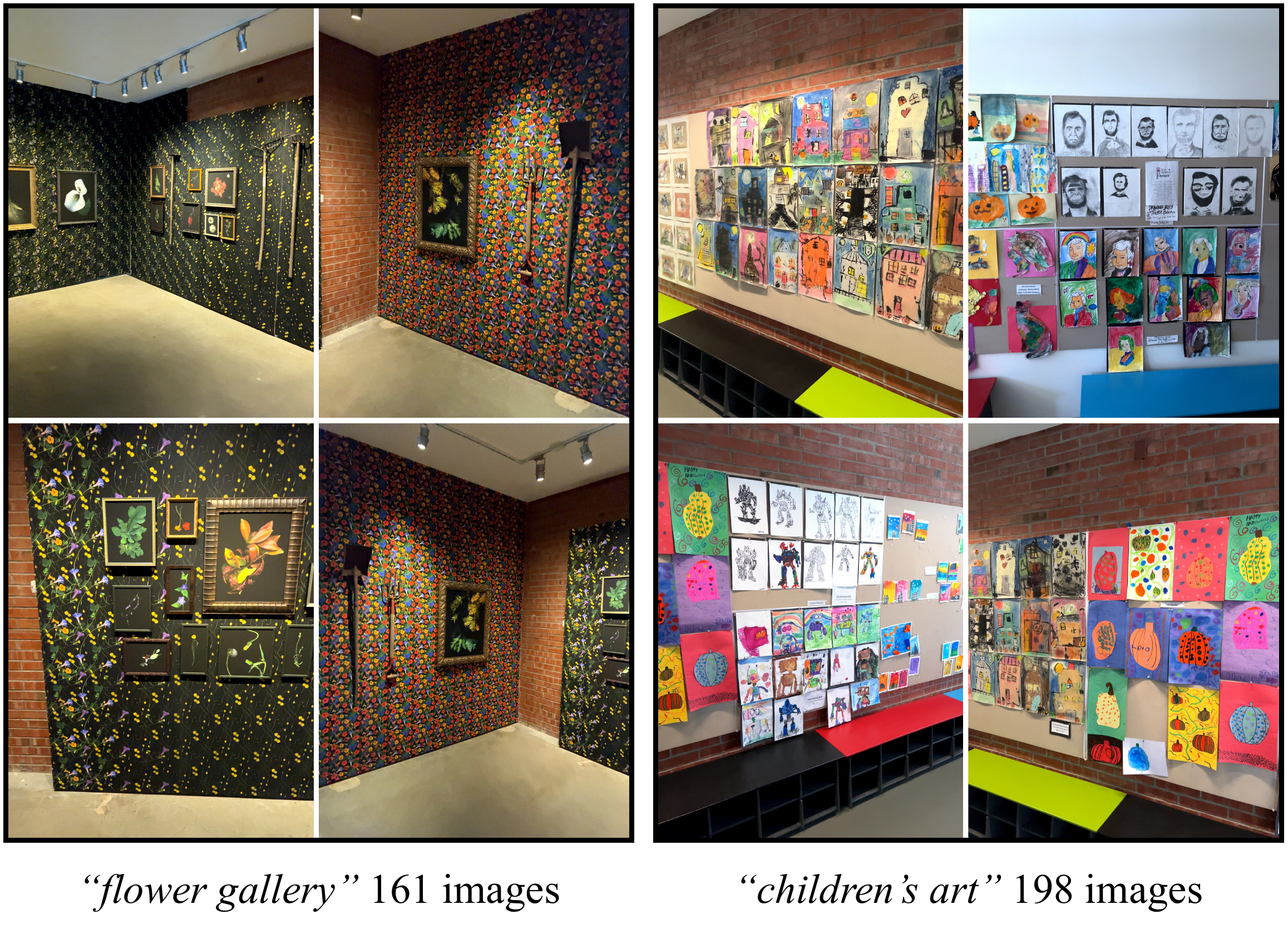}
    \caption{\textbf{Sample training views and number of training images in our custom-captured datasets.} We captured room-level scenes containing highly-detailed artworks to demonstrate the effectiveness of our method. }
    \label{fig:custom_dataset}
\end{figure}

For our custom dataset, we capture two room-scale scenes (\emph{\textquotedblleft flower gallery\textquotedblright} and \emph{\textquotedblleft children's art\textquotedblright}) at a local art center. The images from each dataset are captured by a photographer standing at the center of the room scanning the whole room in a 360 degrees fashion. The number of captured images for each scene in shown in Figure \ref{fig:custom_dataset}.

\noindent \textbf{DTU Dataset Baseline Comparisons Details} For comparisons with Mip-NeRF 360 \cite{barron2022mipnerf360} and Instant-NGP \cite{mueller2022instant} on the DTU dataset, we run the open-source code provided by the authors (\href{https://github.com/google-research/multinerf}{code for Mip-NeRF 360} and \href{https://github.com/NVlabs/instant-ngp}{code for Instant-NGP}). For Mip-NeRF 360 optimization, we slightly modified the configuration file \texttt{multinerf/configs/blender\_256.gin}, which is the default configuration for their Blender experiments. Specifically, we set parameters \texttt{Model.bg\_intensity\_range = (0, 0)} and \texttt{Model.opaque\_background = False} because the DTU dataset scenes all have black backgrounds. The near and far clipping planes are set as $(2, 6)$, respectively, based on the input point cloud and camera positions. The capacity of this model is more than sufficient because DTU dataset has both lower image resolution and less training images than the Blender dataset (one-third to one-fourth of the training images).

For Instant-NGP optimization, we used the configuration file \texttt{instant-ngp/configs/nerf/base.json}, which is the default configuration for all Instant-NGP experiments, including experiments on scene-level datasets.

\section{Additional Quantitative Results}

\noindent \textbf{NVS with Varying Numbers of Gaussians and Fixed Amount of Texels. } We show the quantitative novel-view synthesis results of our RGBA Textured Gaussians models and 3DGS$^*$ with varying numbers of Gaussians in terms of PSNR, SSIM, and LPIPS for the 5 test datasets in Table \ref{tab:supp_same_num_gs} and Figure \ref{fig:supp_num_gaussians_ablation}. We allocate a fixed amount of texels to each of our models. Hence, the texture map resolutions of models with more Gaussians are smaller. Our method consistently outperforms 3DGS$^*$ in terms of PSNR and LPIPS when using different numbers of Gaussians, and the performance improvement is especially large when using fewer Gaussians. 

\noindent \textbf{NVS with Varying Texture Map Resolution and Fixed Number of Gaussians. } We show quantitative novel-view synthesis results of our RGBA Textured Gaussians models with a fixed number of Gaussians and varying texture map resolutions in terms of PSNR, SSIM, and LPIPS in Figure \ref{fig:supp_fix_num_gs_vary_tex_res}. We optimized all models with 1\% of the default optimized number of Gaussians. As expected, the quantitative performance becomes better as the texture resolution increases since the detailed appearances are reconstructed better. 

\noindent \textbf{NVS with Varying Number of Gaussians and Fixed Texture Map Resolution. } We show quantitative novel-view synthesis results of our RGBA Textured Gaussians models with a fixed texture map resolution and varying number of Gaussians in terms of PSNR, SSIM, and LPIPS in Figure \ref{fig:supp_fix_tex_res_vary_num_gs}. We use a $4 \times 4$ texture map for all models. As the number of Gaussians increase, the quantitative performances improve since detailed geometry and appearance are reconstructed better with more, and therefore smaller, Gaussians.

\noindent \textbf{Ablation on Texture Map Variants. } We ablate variants of our model that use different texture maps, namely alpha-only, RGB, and RGBA texture maps, and the same number of Gaussians. Experiments are conducted on all five standard benchmark datasets with the default optimized number (top row) and 1\% of the default optimized number (bottom row) of Gaussians. We report the PSNR/SSIM/LPIPS values of the novel-view synthesis results. From Table \ref{tab:supp_tex_ablation} and Figure \ref{fig:supp_texture_map_ablation}, we see that our model achieves the best performance with RGBA textures. Interestingly, using alpha-only textures already outperforms 3DGS$^*$ and our RGB textures model despite being one-third the size, striking a better balance between performance and model size. This is because alpha-textured Gaussians can represent complex shapes through spatially varying opacity and reconstruct high-frequency textures through spatially varying alpha composition. In contrast, RGB-textured Gaussians are still limited to representing ellipsoids.

\noindent \textbf{Computational Efficiency Analysis. } The total training time of Textured Gaussians is $2\times$ that of 3DGS due to the two-stage training process. In Table \ref{tab:efficiency}, we report various efficiency metrics, aggregated over all 26 test scenes, of our Textured Gaussians vs.\ 3DGS using the same number of Gaussians and the setup of Table 1 in the main paper (100\% of the default number of optimized Gaussians and $4\times4$ texture maps). 

As for the other texture map variants of Textured Gaussians, the model sizes of using alpha-only, RGB, and RGBA textures are $1.27\times, 1.82\times, 2.08\times$ that of 3DGS with the same number of Gaussians.
Note that using alpha-only textures only imposes slight memory overhead and performs better than 3DGS, according to Figure 8 in the paper.

\begin{table}[!t]
    \centering
    \footnotesize
    \begin{tabular*}{\linewidth}{@{\extracolsep{\fill}}l@{\hskip 0.1cm}c@{\hskip 0.1cm}c@{}}
        \toprule
        Model & 3DGS & Ours (w/ 4$\times$4 RGBA textures)\\
        \midrule
        Number of Gaussians (M) & 2.12 M & 2.12 M  \\
        Model size (MB) & 125.3 MB & 261.2 MB \\
        GPU memory (GB) & 2.4 & 4.0 \\
        Inference time per frame (s) & 0.073 & 0.123 \\
        \bottomrule
    \end{tabular*}
    \caption{We report various efficiency metrics of 3DGS and our Textured Gaussians model. With the same number of Gaussians, the model size and GPU usage of Textured Gaussians are around $2\times$ that of 3DGS. The inference speed is also slower due to additional textured map queries. }
    \label{tab:efficiency}
\end{table}

\section{Additional Qualitative Results} 
In this section, we show qualitative NVS results and refer the readers to the \href{https://textured-gaussians.github.io/}{project website} for novel-view \textbf{video} renderings of selected scenes for each experiment. 

\noindent \textbf{NVS with Varying Numbers of Gaussians and Fixed Amount of Texels. } We show novel-view synthesis results from the 24 test scenes that we used for evaluation in Figures \ref{fig:extra_nvs_blender} to \ref{fig:extra_nvs_db}. We show results of 3DGS$^*$ and our Textured Gaussians model using the default optimized number and 1\% of the default optimized number of Gaussians. Our method achieves sharper details than 3DGS$^*$ in both cases. The differences between the two methods are especially clear when using fewer Gaussians, as 3DGS$^*$ struggles to reconstruct fine-grained textures while our method achieves decent image quality.

We additionally show novel-view video renderings of selected scenes optimized using 3DGS$^*$ and our RGBA Textued Gaussians model with varying number of Gaussians (continuously varying from 1\% to 100\% of the default number of optimized Gaussians) in the \href{https://textured-gaussians.github.io/}{project website}. We highly recommend readers to try out the interactive website and toggle between models optimized with different numbers of Gaussians to see how novel-view synthesis quality improves as the number of Gaussians increases.

\noindent \textbf{NVS with Varying Texture Map Resolution and Fixed Number of Gaussians. } We show qualitative novel-view synthesis results of our RGBA Textured Gaussians models with a fixed number of Gaussians and varying texture map resolutions in Figures \ref{fig:vary_tex_res_blender} to \ref{fig:vary_tex_res_db}. Due to space constraints, we show results of our Textured Gaussians models with texture map resolutions $\mathcal{T} = 2, 5, 20, 40$.

As expected, NVS performance improves as the texture map resolution increases, since detailed appearances can be reconstructed better with smaller texture feature sizes.

\noindent \textbf{NVS with Varying Number of Gaussians and Fixed Texture Map Resolution. } We show qualitative novel-view synthesis results of our RGBA Textured Gaussians models with a fixed texture map resolution and varying numbers of Gaussians in Figures \ref{fig:vary_num_gs_blender} to \ref{fig:vary_num_gs_db}. Due to space constraints, we show results of our Textured Gaussians models 1\%, 5\%, 20\%, and 100\% of the default number of optimized Gaussians.

As expected, NVS performance improves as the number of Gaussians increase since details can be reconstructed better with more and therefore, smaller, Gaussians.

\noindent \textbf{Ablation on Texture Map Variants. } We show novel view synthesis results of our Textured Gaussians model optimized using different texture map variants, namely alpha-only, RGB, and RGBA textures, in Figures \ref{fig:tex_ablation_blender} to \ref{fig:tex_ablation_db}. Here, we show the results of our models and 3DGS$^*$ with 1\% of the default optimized number of Gaussians to maximize the visual difference between different variants to fully demonstrate the effectiveness of our method.  

From the results, we see that the alpha-textures model and the RGBA textures model achieve better qualitative performance than the RGB textures model. This is because alpha-textured Gaussians are able to both reconstruct fine-grained textures through spatially varying alpha composition and represent complex shapes through spatially varying opacity. On the other hand, RGB-textured Gaussians are still only able to represent ellipsoids. This is clearly observed in scenes that contain complex geometric structures, such as the \textit{ship} scene in the Blender dataset (last row) and the lego truck in the \textit{kitchen} scene in the mipNeRF360 dataset (6th row).

We additionally show novel-view video renderings of selected scenes optimized using 3DGS$^*$ and different variants (alpha, RGB, and RGBA) of our Textured Gaussians model in the \href{https://textured-gaussians.github.io/}{project website}. We highly recommend readers to try out the interactive website and toggle between models optimized with different texture map variants to see how alpha and RGB texture maps affect novel-view rendering quality.

\noindent \textbf{Color Component Decompositions. } We optimized our Textured Gaussians model with alpha-only, RGB, and RGBA textures with the same number of Gaussians and show the color decomposition of the final rendered color into the two color components $\mathbf{c}_\text{base}$ and $\mathbf{c}_\text{tex}$ in Figures \ref{fig:color_contribution_blender} to \ref{fig:color_contribution_db}. Specifically, we show the \textit{alpha-modulated and composited} colors of the color components. 3DGS$^*$ models optimized with the same number of Gaussians are also shown for better comparison. We optimize all models with 1\% of the default optimized number of Gaussians to maximize the visual difference between our method and 3DGS$^*$ to fully demonstrate the effectiveness of our method. 

We see from the results of the alpha-only and RGBA textures model that although the color of $\mathbf{c}_\text{base}$ comes from spatially constant spherical harmonic coefficients, the alpha-modulated $\mathbf{c}_\text{base}$ is able to reconstruct high-frequency textures due to the spatially varying alpha composition. On the other hand, $\mathbf{c}_\text{base}$ in the RGB textures model cannot reconstruct high-frequency textures, and $\mathbf{c}_\text{tex}$ is used to reconstruct detailed appearance.

We additionally show novel-view video renderings of the color component decompositions of selected scenes optimized using 3DGS$^*$ and different variants (alpha, RGB, and RGBA) of our Textured Gaussians model in the \href{https://textured-gaussians.github.io/}{project website}. We highly recommend readers to try out the interactive website and toggle between models optimized with different texture map variants to see how the color component decompositions differ from one another.

\begin{figure}[!t]
    \centering
    \includegraphics[width=\linewidth]{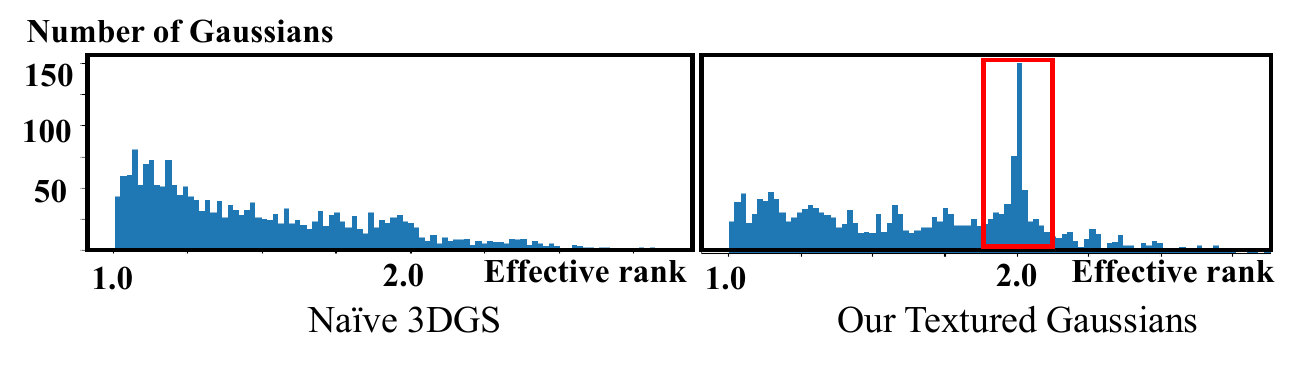}
    \caption{We analyzed the effective rank distribution \cite{hyung2024erank} of na\"ive Gaussian splatting and Textured Gaussians models. Our Textured Gaussians optimization naturally leads to a large proportion of flat 2D Gaussians (effective rank = 2), while na\"ive Gaussian splatting leads to more needle-like Gaussian strands (effective rank $\approx 1$). This observation justifies our choice of using the 3DGS framework  to approximate 2D Gaussians.}
    \label{fig:eff_rank}
\end{figure}

\noindent \textbf{3DGS vs.\ 2DGS-based implementation. } We find that Textured Gaussians optimization naturally leads to a large proportion of flat Gaussians, as the distributions of effective rank are shown Figure \ref{fig:eff_rank}. Note that flat 3D Gaussians have an effective rank of 2 \cite{hyung2024erank}. This observation justifies our design choice to flatten 3D Gaussians to approximate 2D Gaussians, similar to prior works \cite{Yu2024GOF, guedon2023sugar}.

\begin{figure}[!t]
    \centering
    \includegraphics[width=\linewidth]{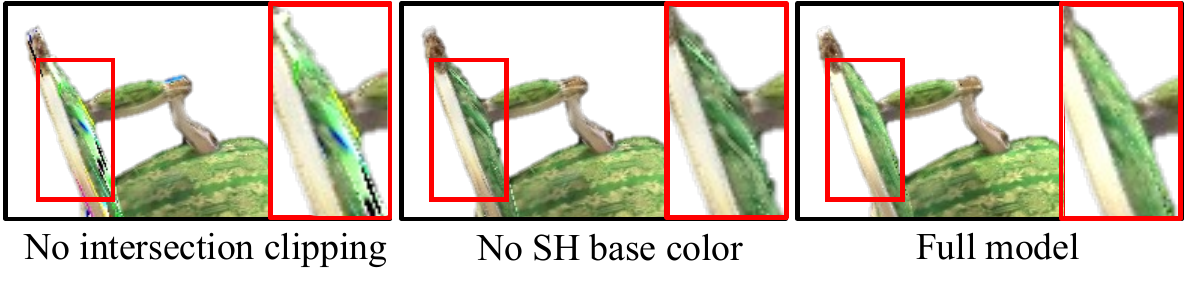}
    \caption{We designed various heuristics to reduce artifacts caused by rendering flat 2D Gaussians from grazing viewpoints, including intersection clipping (left) and adding SH base colors (middle). Our full model achieves the best rendering results.}
    \label{fig:intersection_ablation}
\end{figure}

\begin{figure*}
    \centering
    \includegraphics[width=\linewidth]{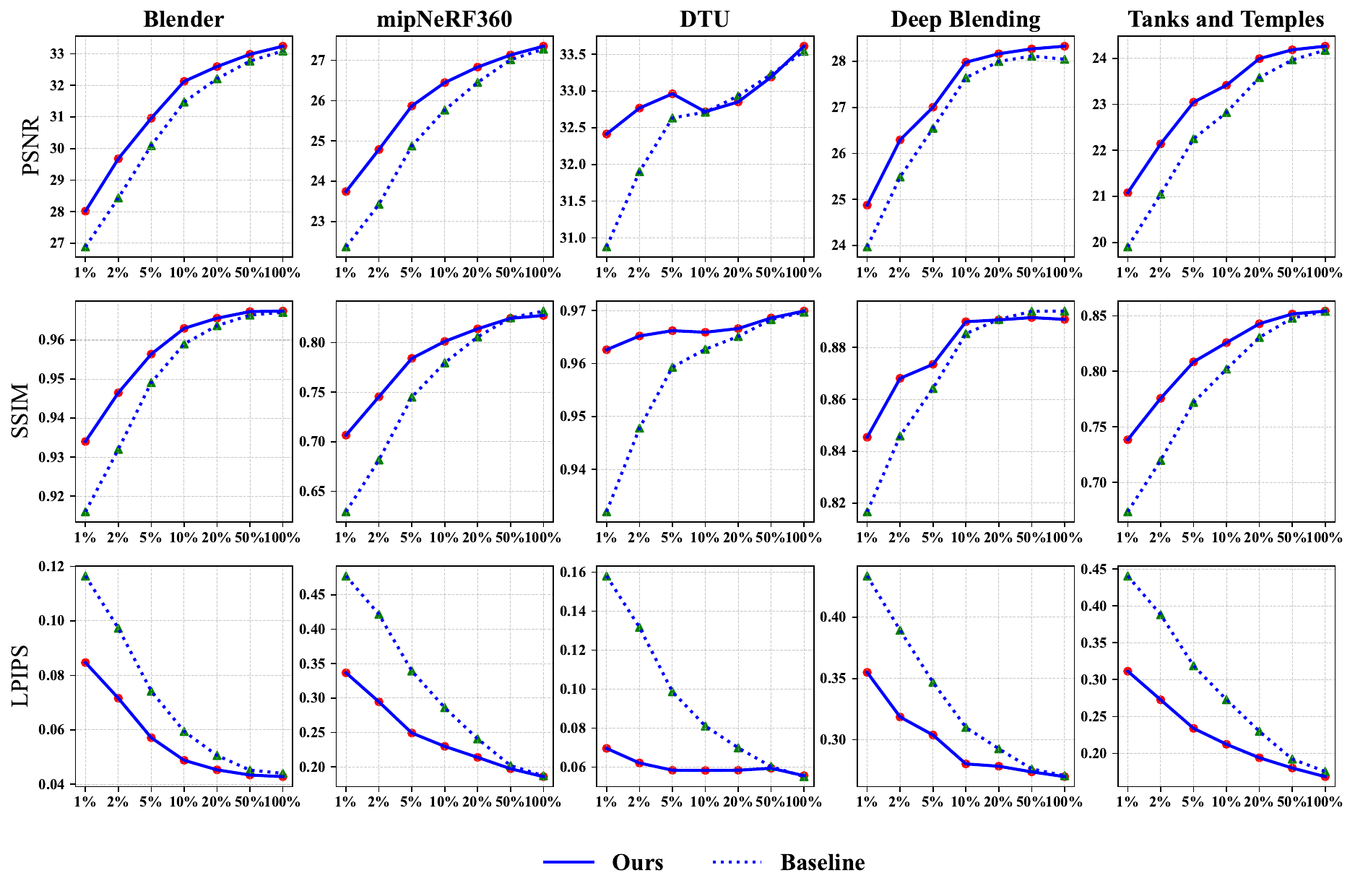}
    \caption{\textbf{Quantitaive results on all datasets using varying numbers of Gaussians with a fixed amount of texels. } We show the NVS performance trend of our RGBA Textured Gaussians model and 3DGS$^*$ as the number of Gaussians increases. Our Textured Gaussians model greatly outperforms 3DGS$^*$ with fewer Gaussians, and the performance gap between the two methods decreases as the number of Gaussians increases.}
    \label{fig:supp_num_gaussians_ablation}
\end{figure*}

\begin{figure*}
    \centering
    \includegraphics[width=\linewidth]{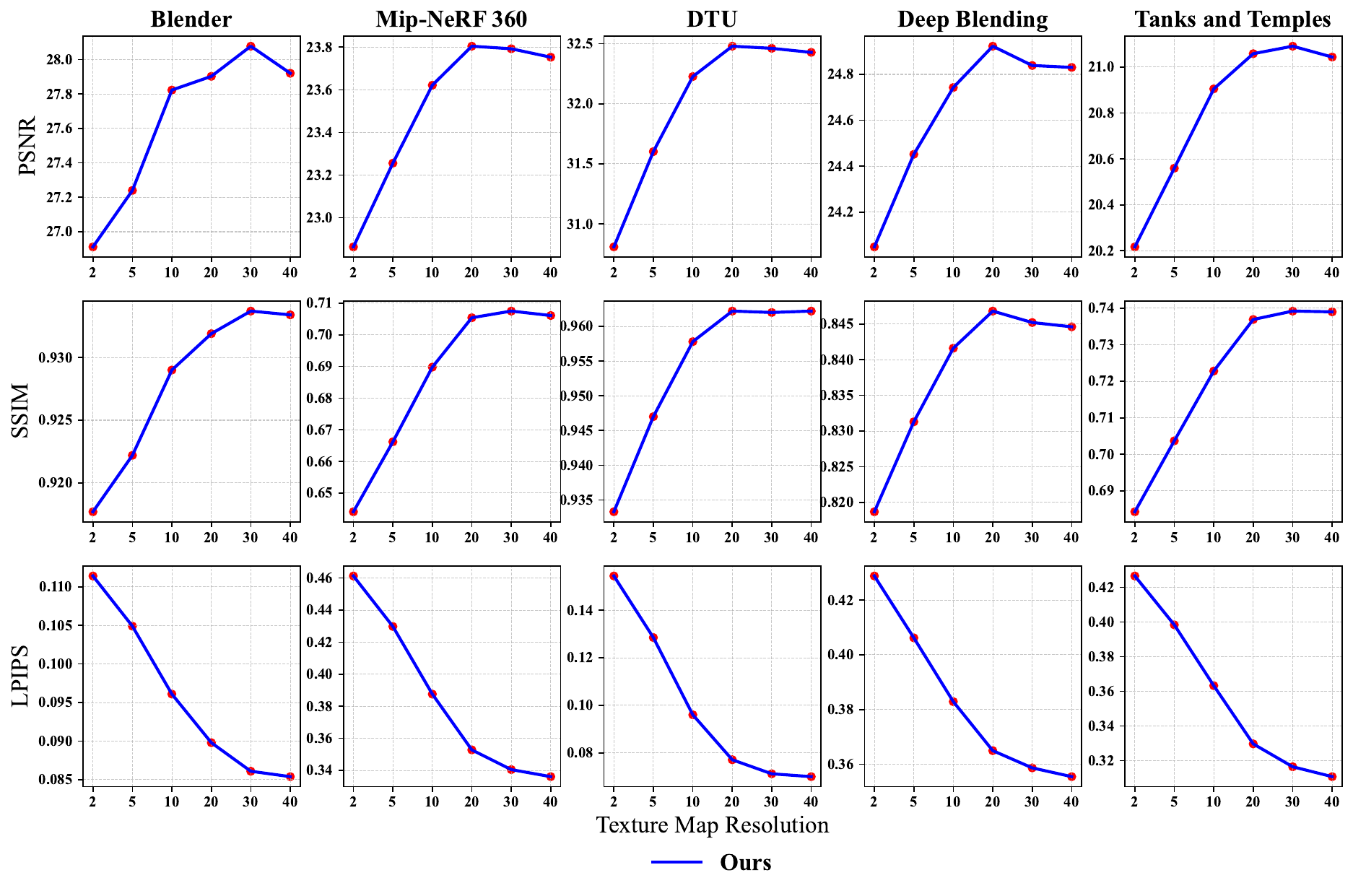}
    \caption{\textbf{Quantitative results of ablation study on varying texture resolutions with a fixed number of Gaussians. } As the texture map resolution increases, novel-view synthesis quality improves as detailed appearances are reconstructed better. However, the performance drops slightly when the texture map resolution is too high, likely due to overfitting. }
    \label{fig:supp_fix_num_gs_vary_tex_res}
\end{figure*}

\begin{figure*}
    \centering
    \includegraphics[width=\linewidth]{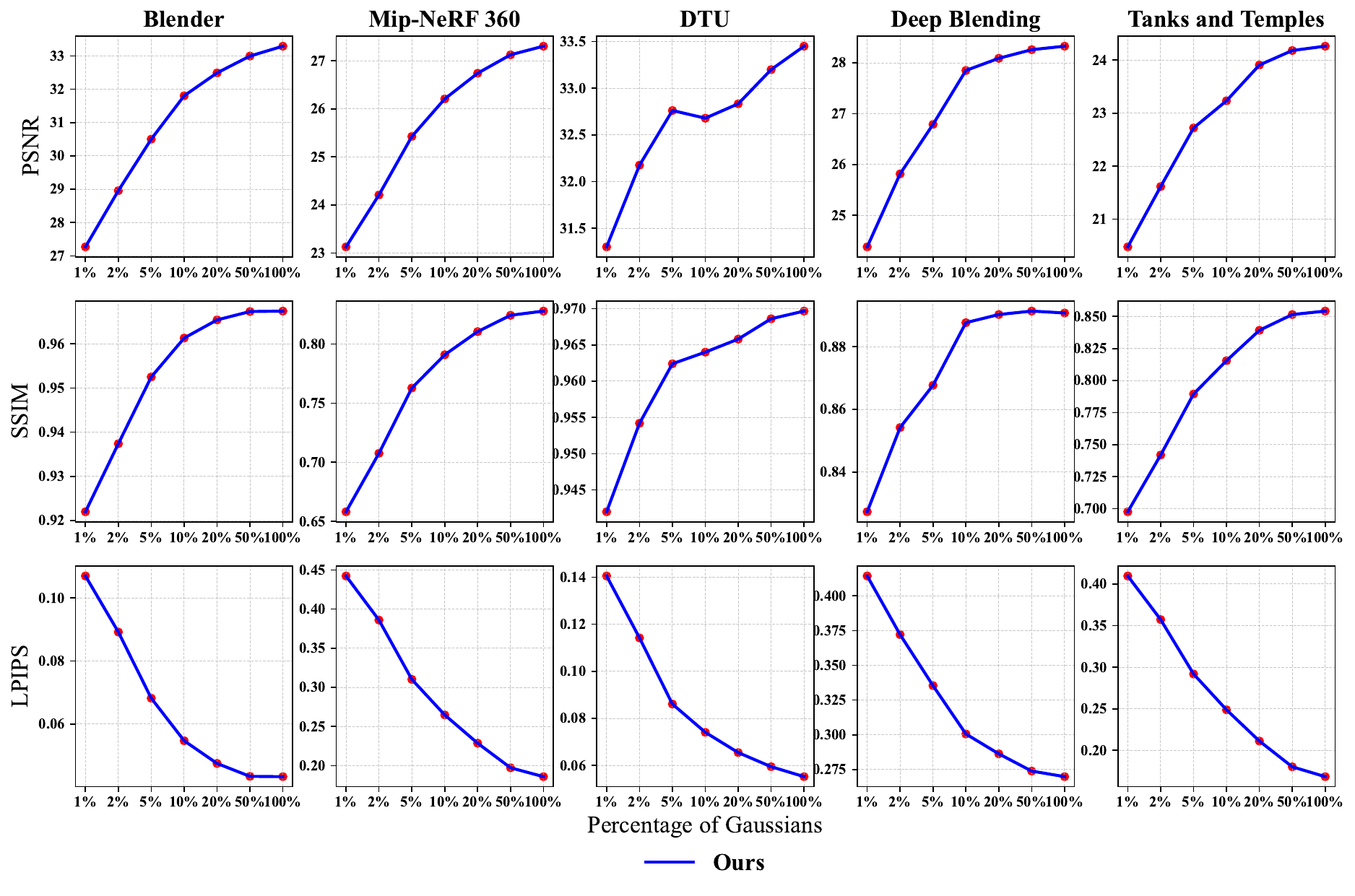}
    \caption{\textbf{Quantitative results of ablation study on varying number of Gaussians and with a fixed texture map resolution. } As the number of Gaussians increase, novel-view synthesis performance improves since detailed appearance and geometry can be reconstructed better using smaller and more Gaussians.}
    \label{fig:supp_fix_tex_res_vary_num_gs}
\end{figure*}

\begin{figure*}
    \centering
    \includegraphics[width=\linewidth]{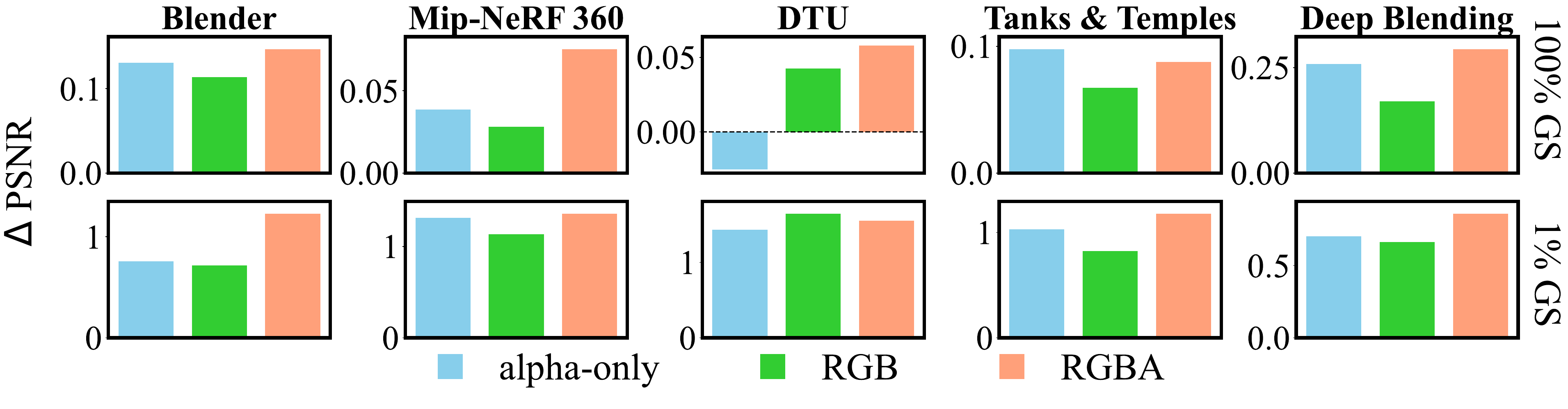}
    \caption{\textbf{Quantitative results of ablation study on texture map variants. } We optimized our Textured Guasians model with different texture map variants (alpha-only, RGB, RGBA). We show the difference in PSNR values compared to the 3DGS$^*$ baseline for better visual comparison. We observe that the alpha-only textures models generally achieve better performance than RGB textures models. This is because alpha textures are capable of representing complex appearance and shapes due to spatially varying alpha modulation and composition. On the other hand, RGB-textured Gaussians are still only capable of representing ellipsoids. The full RGBA textures models achieve the best results since the RGBA-textured Gaussians are most expressive. }
    \label{fig:supp_texture_map_ablation}
\end{figure*}

\begin{figure*}
    \centering
    \includegraphics[width=\linewidth]{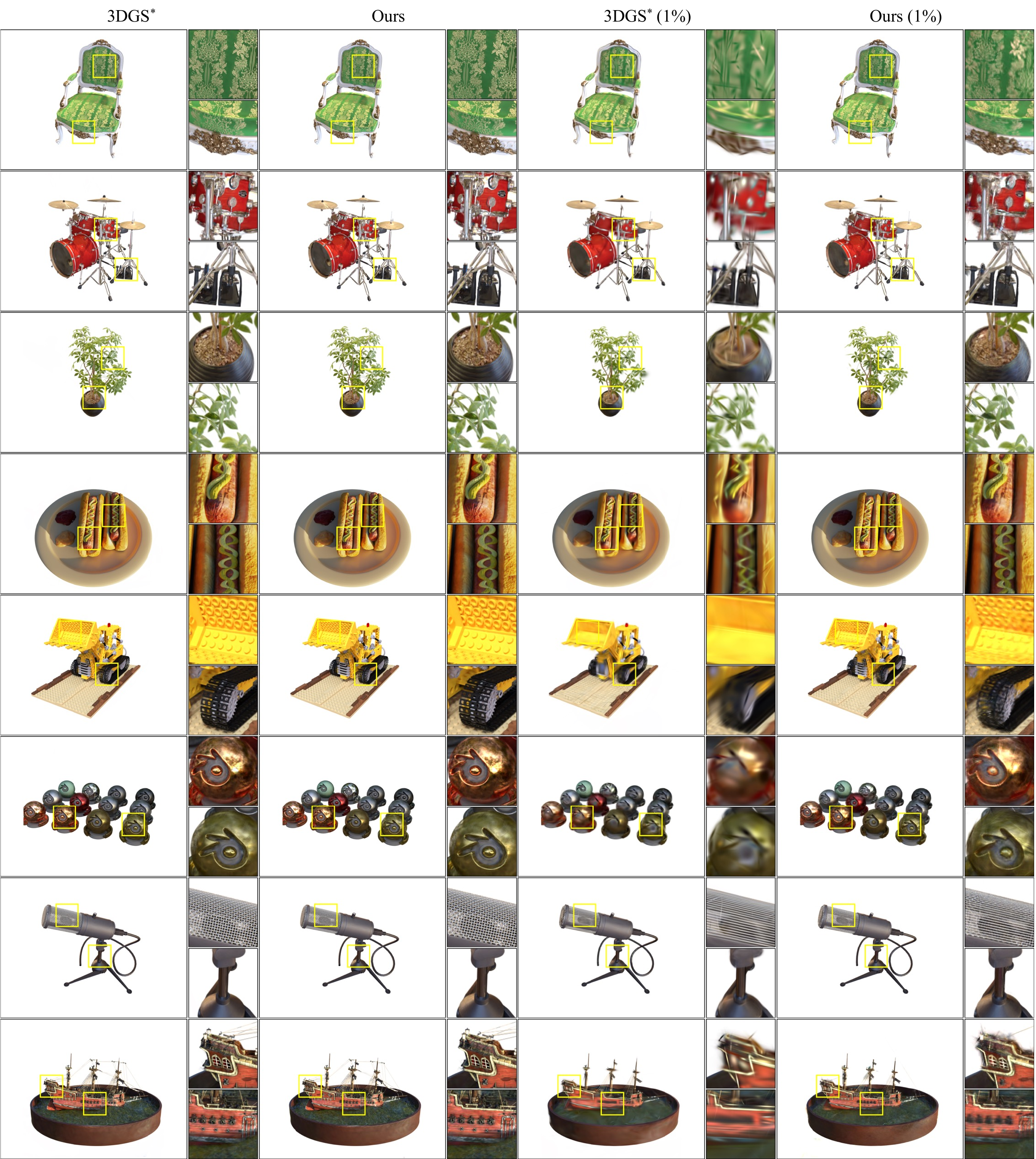}
    \caption{\textbf{Qualitative NVS results on the Blender dataset with varying numbers of Gaussians and a fixed amount of texels.} Our RGBA Textured Gaussians model achieves better NVS quality compared to 3DGS$^*$ when using the same number of Gaussians. The visual difference is especially clear with fewer Gaussians. }
    \label{fig:extra_nvs_blender}
\end{figure*}

\begin{figure*}
    \centering
    \includegraphics[height=0.95 \textheight]{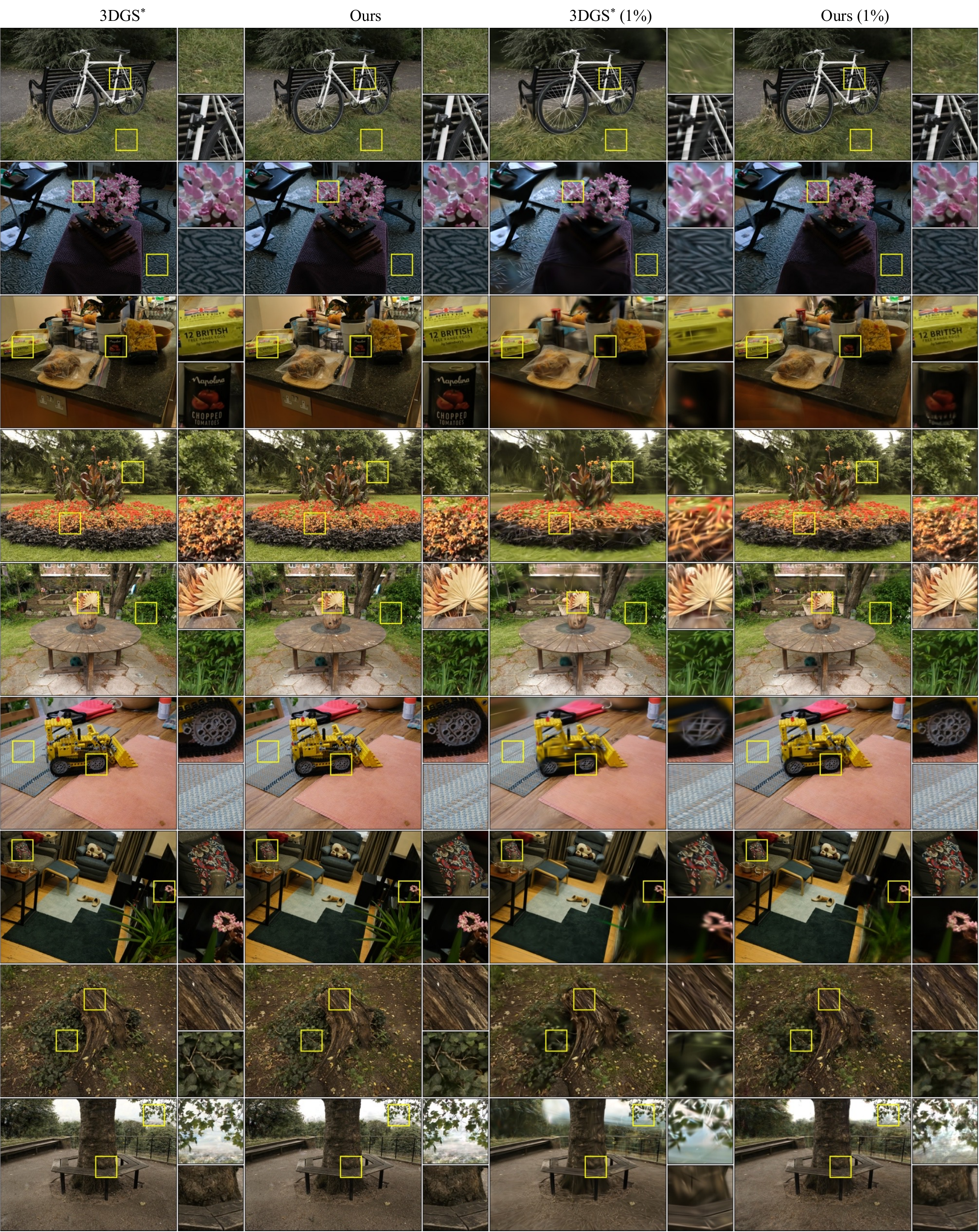}
    \caption{\textbf{Qualitative NVS results on the Mip-NeRF 360 dataset with varying numbers of Gaussians and a fixed amount of texels.} Our RGBA Textured Gaussians model achieves better NVS quality compared to 3DGS$^*$ when using the same number of Gaussians. The visual difference is especially clear with fewer Gaussians. }
    \label{fig:extra_nvs_mipnerf360}
\end{figure*}

\begin{figure*}
    \centering
    \includegraphics[width=\linewidth]{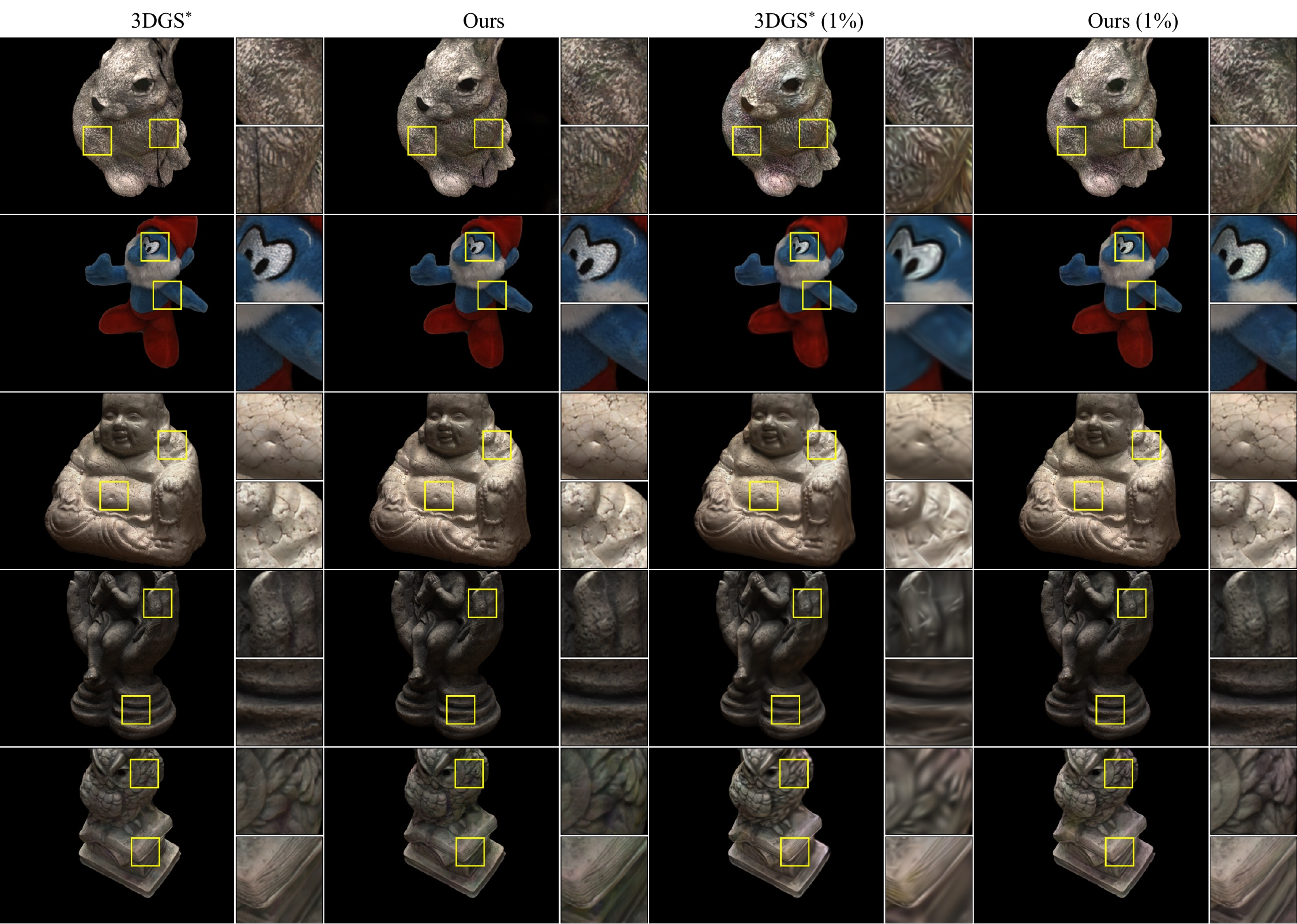}
    \caption{\textbf{Qualitative NVS results on the DTU dataset with varying numbers of Gaussians and a fixed amount of texels.} Our RGBA Textured Gaussians model achieves better NVS quality compared to 3DGS$^*$ when using the same number of Gaussians. The visual difference is especially clear with fewer Gaussians. }
    \label{fig:extra_nvs_dtu}
\end{figure*}

\begin{figure*}
    \centering
    \includegraphics[width=\linewidth]{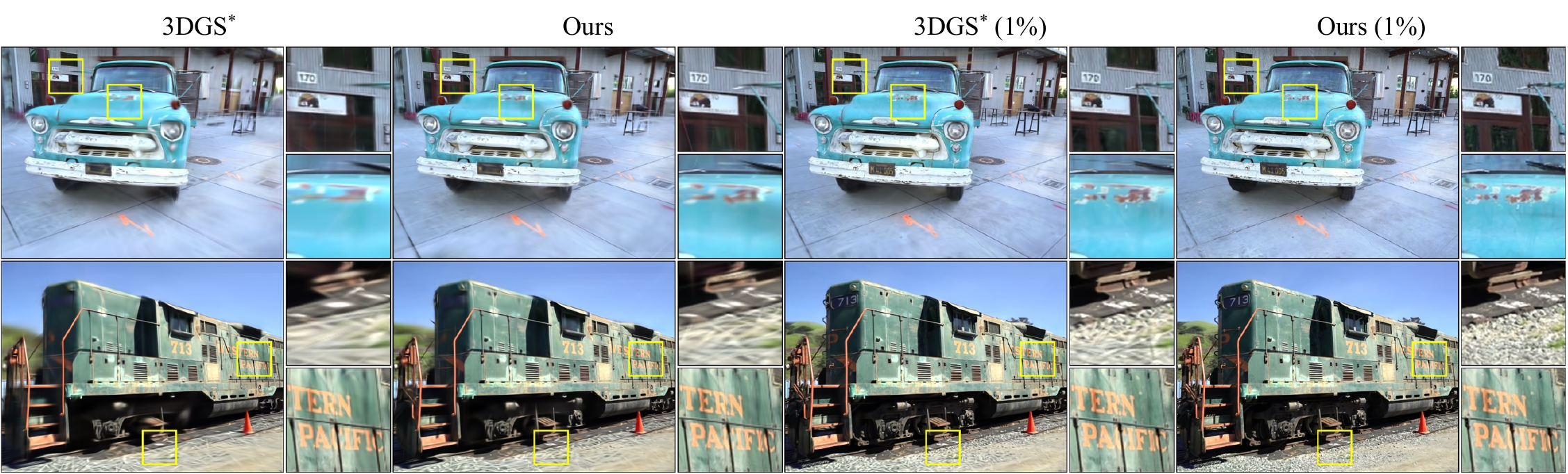}
    \caption{\textbf{Qualitative NVS results on the Tanks and Temples dataset with varying numbers of Gaussians and a fixed amount of texels.} Our RGBA Textured Gaussians model achieves better NVS quality compared to 3DGS$^*$ when using the same number of Gaussians. The visual difference is especially clear with fewer Gaussians. }
    \label{fig:extra_nvs_tandt}
\end{figure*}

\begin{figure*}
    \centering
    \includegraphics[width=\linewidth]{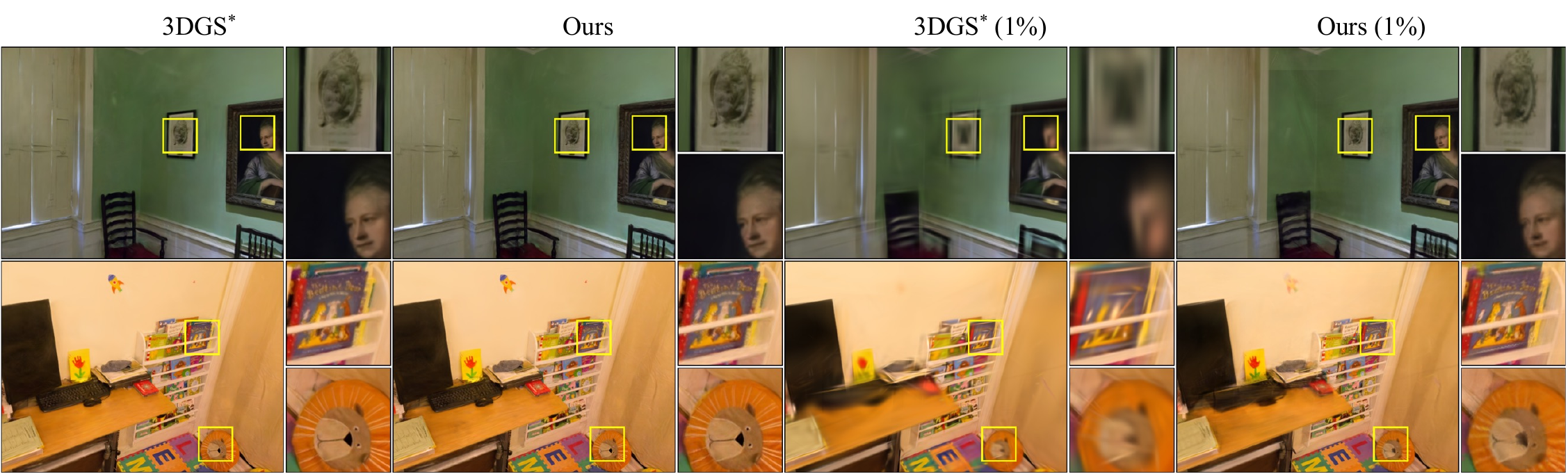}
    \caption{\textbf{Qualitative NVS results on the Deep Blending dataset with varying numbers of Gaussians and a fixed amount of texels.} Our RGBA Textured Gaussians model achieves better NVS quality compared to 3DGS$^*$ when using the same number of Gaussians. The visual difference is especially clear with fewer Gaussians. }
    \label{fig:extra_nvs_db}
\end{figure*}

\begin{figure*}
    \centering
    \includegraphics[width=\linewidth]{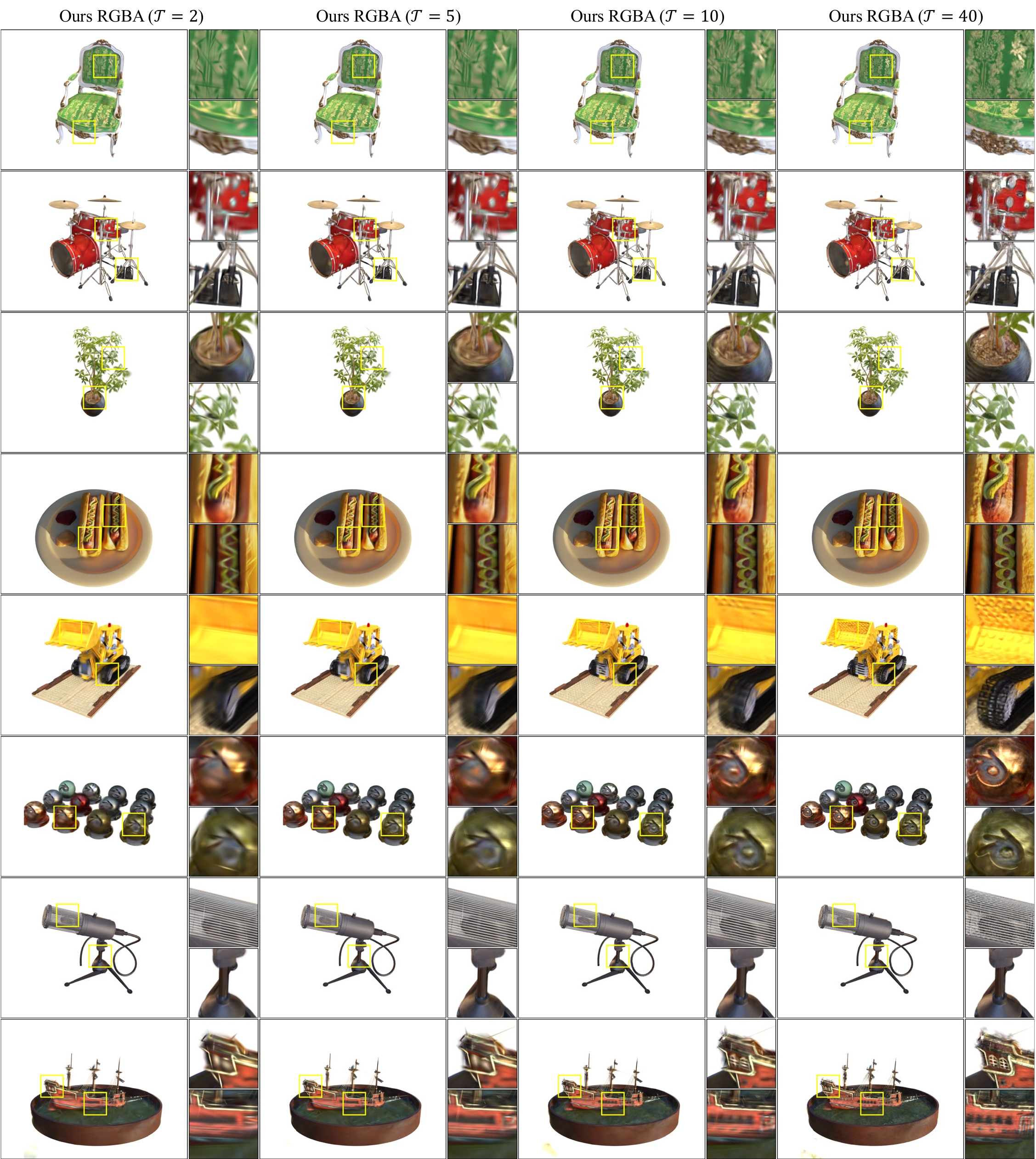}
    \caption{\textbf{Qualitative NVS results on the Blender dataset with vayring texture map resolutions and a fixed number of Gaussians. } Given a fixed number of Gaussians, novel-view synthesis performance improves as the texture map resolution increases. Detailed appearance can be reconstructed better with higher-resolution texture maps, or smaller texel feature sizes. }
    \label{fig:vary_tex_res_blender}
\end{figure*}

\begin{figure*}
    \centering
    \includegraphics[height=0.9\textheight]{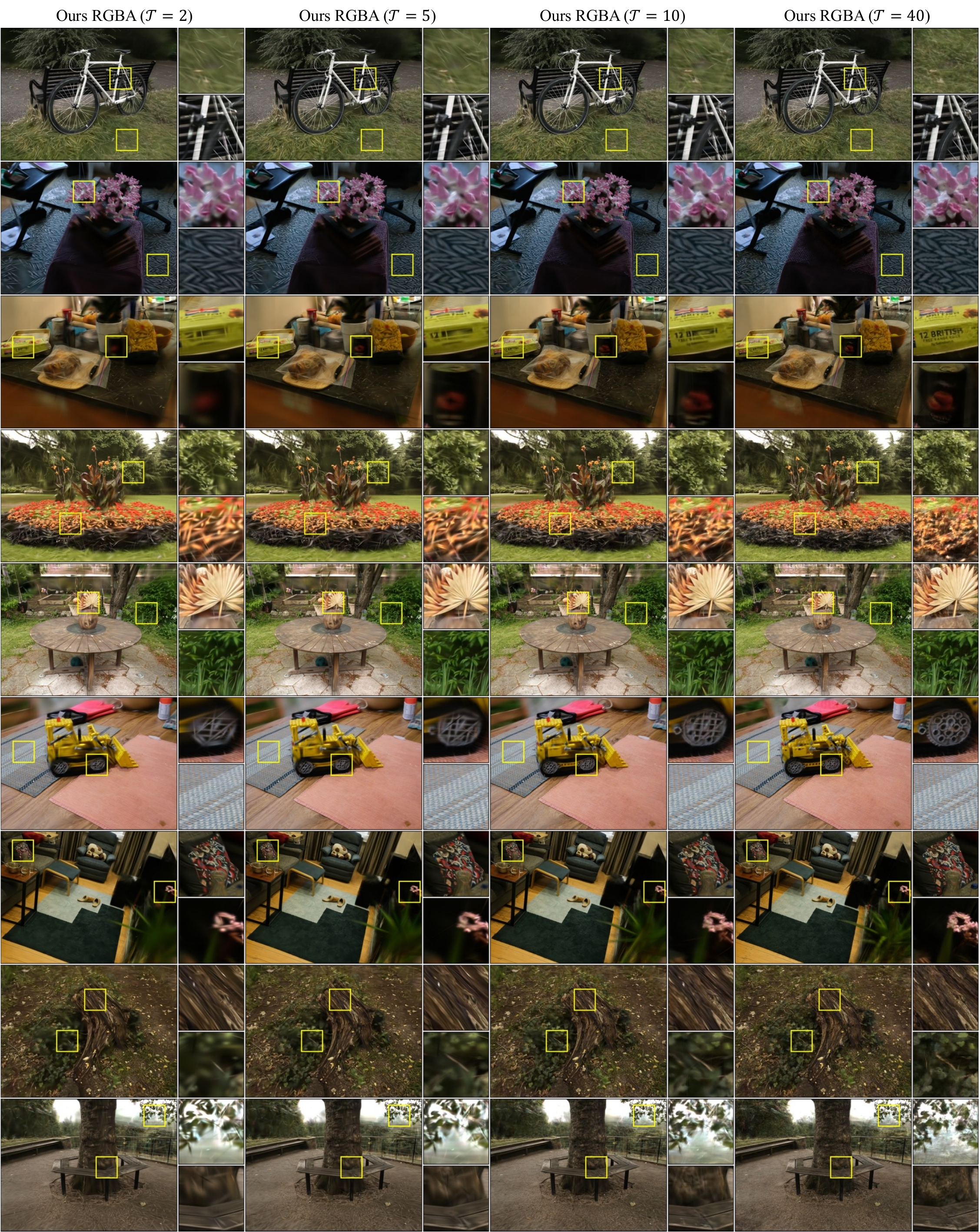}
    \caption{\textbf{Qualitative NVS results on the Mip-NeRF 360 dataset with vayring texture map resolutions and a fixed number of Gaussians. } Given a fixed number of Gaussians, novel-view synthesis performance improves as the texture map resolution increases. Detailed appearance can be reconstructed better with higher-resolution texture maps, or smaller texel feature sizes. }
    \label{fig:vary_tex_res_mipnerf360}
\end{figure*}

\begin{figure*}
    \centering
    \includegraphics[width=\linewidth]{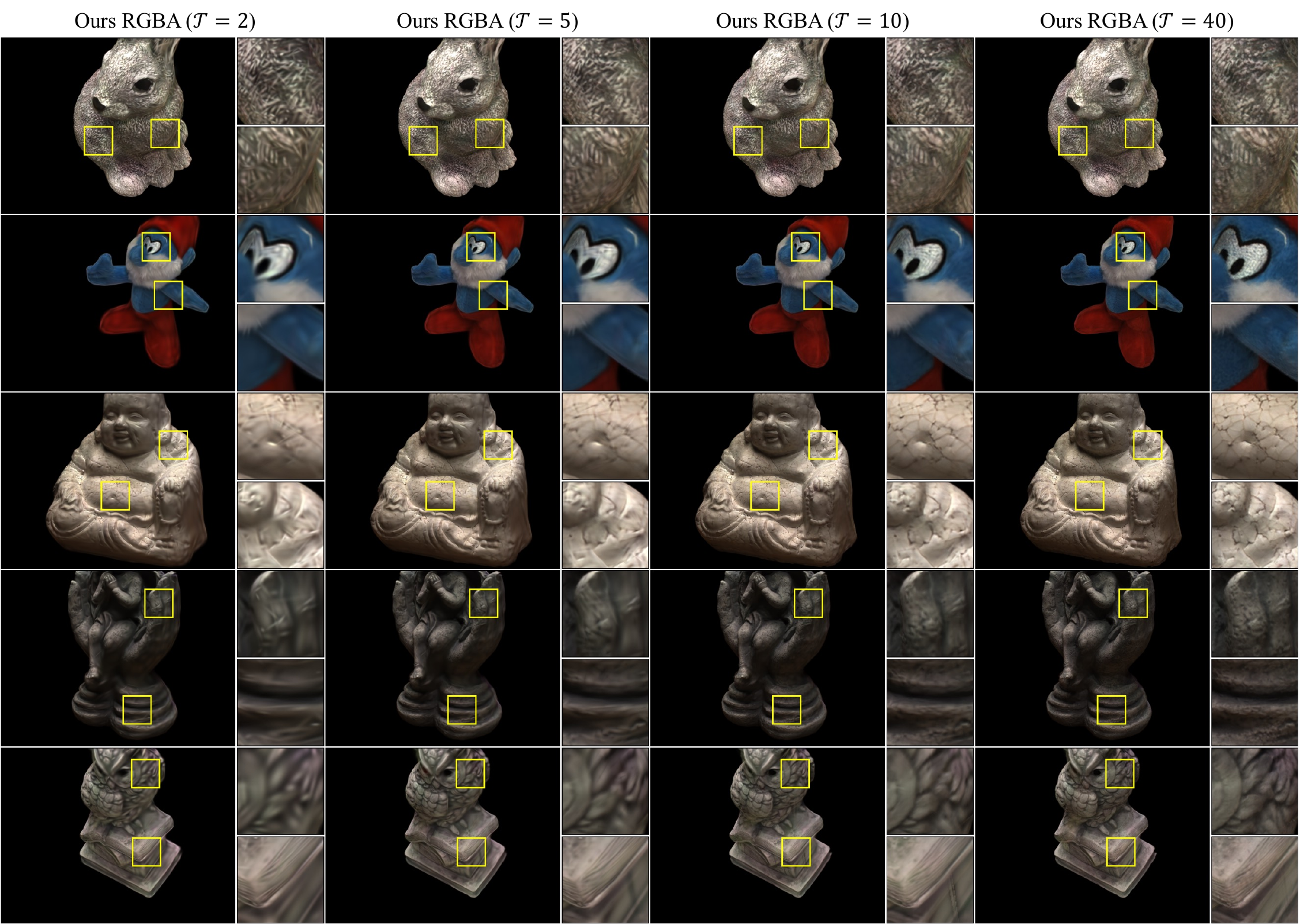}
    \caption{\textbf{Qualitative NVS results on the DTU dataset with vayring texture map resolutions and a fixed number of Gaussians. } Given a fixed number of Gaussians, novel-view synthesis performance improves as the texture map resolution increases. Detailed appearance can be reconstructed better with higher-resolution texture maps, or smaller texel feature sizes. }
    \label{fig:vary_tex_res_dtu}
\end{figure*}

\begin{figure*}
    \centering
    \includegraphics[width=\linewidth]{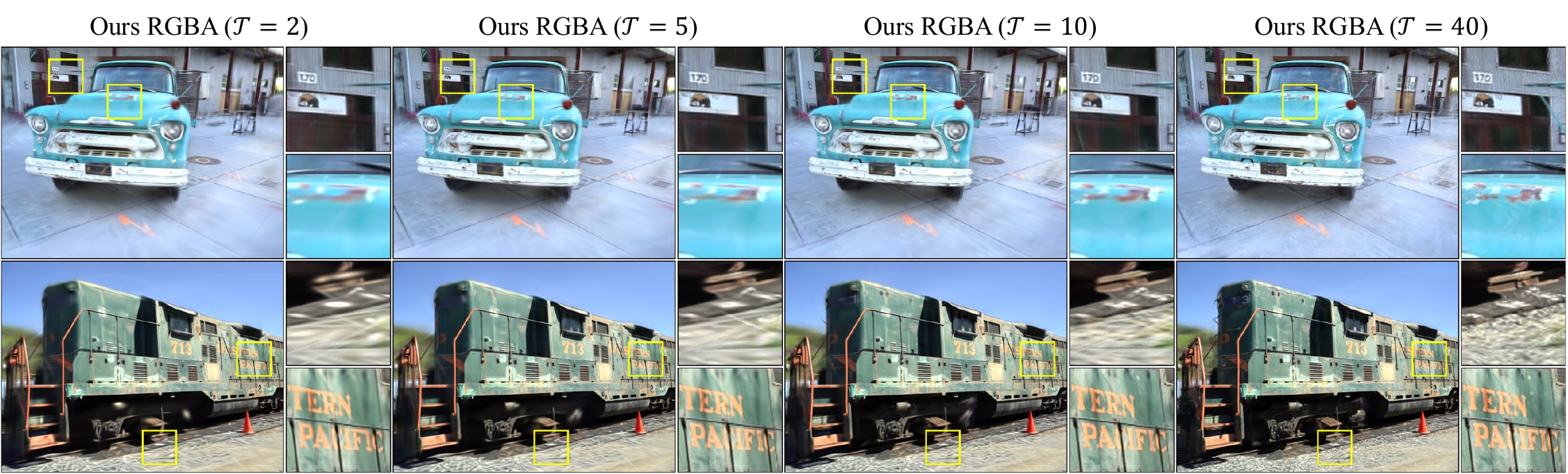}
    \caption{\textbf{Qualitative NVS results on the Tanks and Temples dataset with vayring texture map resolutions and a fixed number of Gaussians. } Given a fixed number of Gaussians, novel-view synthesis performance improves as the texture map resolution increases. Detailed appearance can be reconstructed better with higher-resolution texture maps, or smaller texel feature sizes. }
    \label{fig:vary_tex_res_tandt}
\end{figure*}

\begin{figure*}
    \centering
    \includegraphics[width=\linewidth]{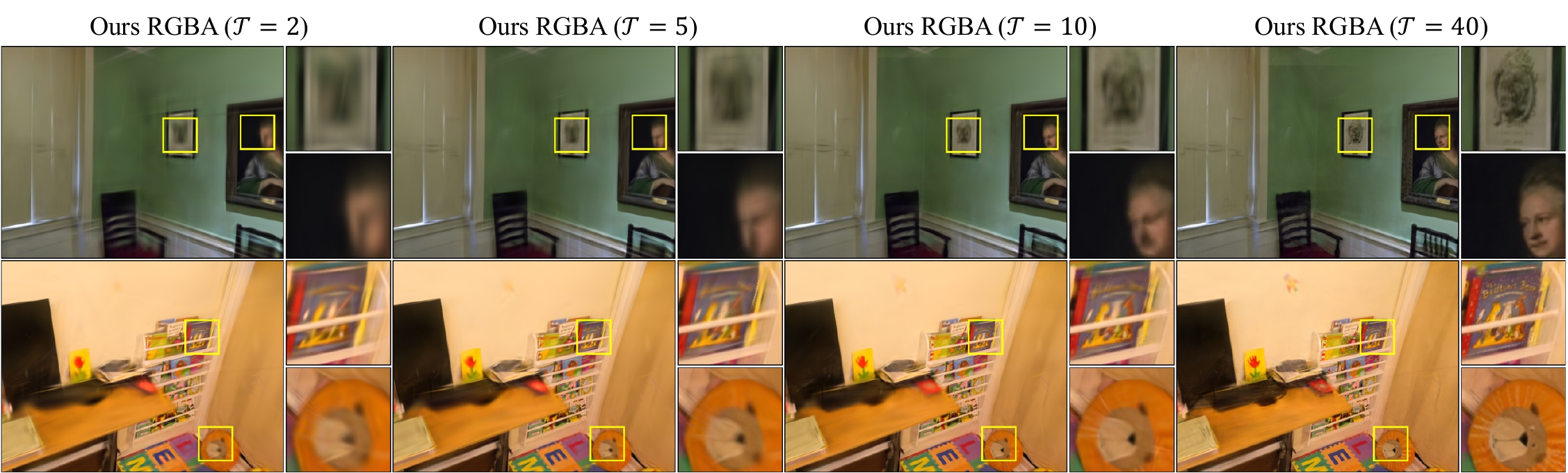}
    \caption{\textbf{Qualitative NVS results on the Deep Blending dataset with vayring texture map resolutions and a fixed number of Gaussians. } Given a fixed number of Gaussians, novel-view synthesis performance improves as the texture map resolution increases. Detailed appearance can be reconstructed better with higher-resolution texture maps, or smaller texel feature sizes. }
    \label{fig:vary_tex_res_db}
\end{figure*}

\begin{figure*}
    \centering
    \includegraphics[width=\linewidth]{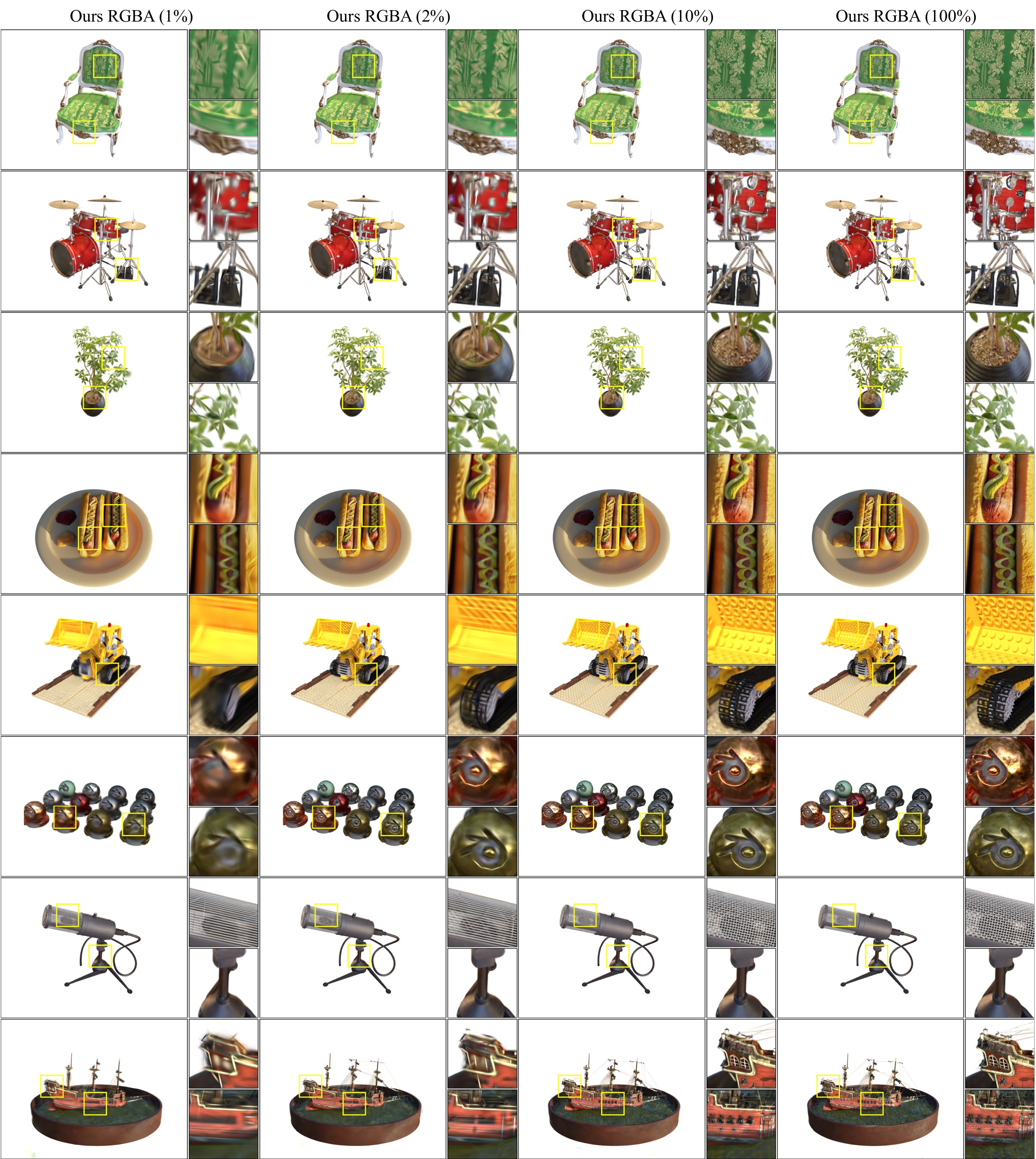}
    \caption{\textbf{Qualitative NVS results on the Blender dataset with vayring numbers of Gaussians and a fixed texture map resolution. } Given a fixed texture map resolution, novel-view synthesis performance improves as the number of Gaussians increases. Detailed appearance and complex geometry can be reconstructed with more and therefore smaller Gaussians.}
    \label{fig:vary_num_gs_blender}
\end{figure*}

\begin{figure*}
    \centering
    \includegraphics[height=0.9\textheight]{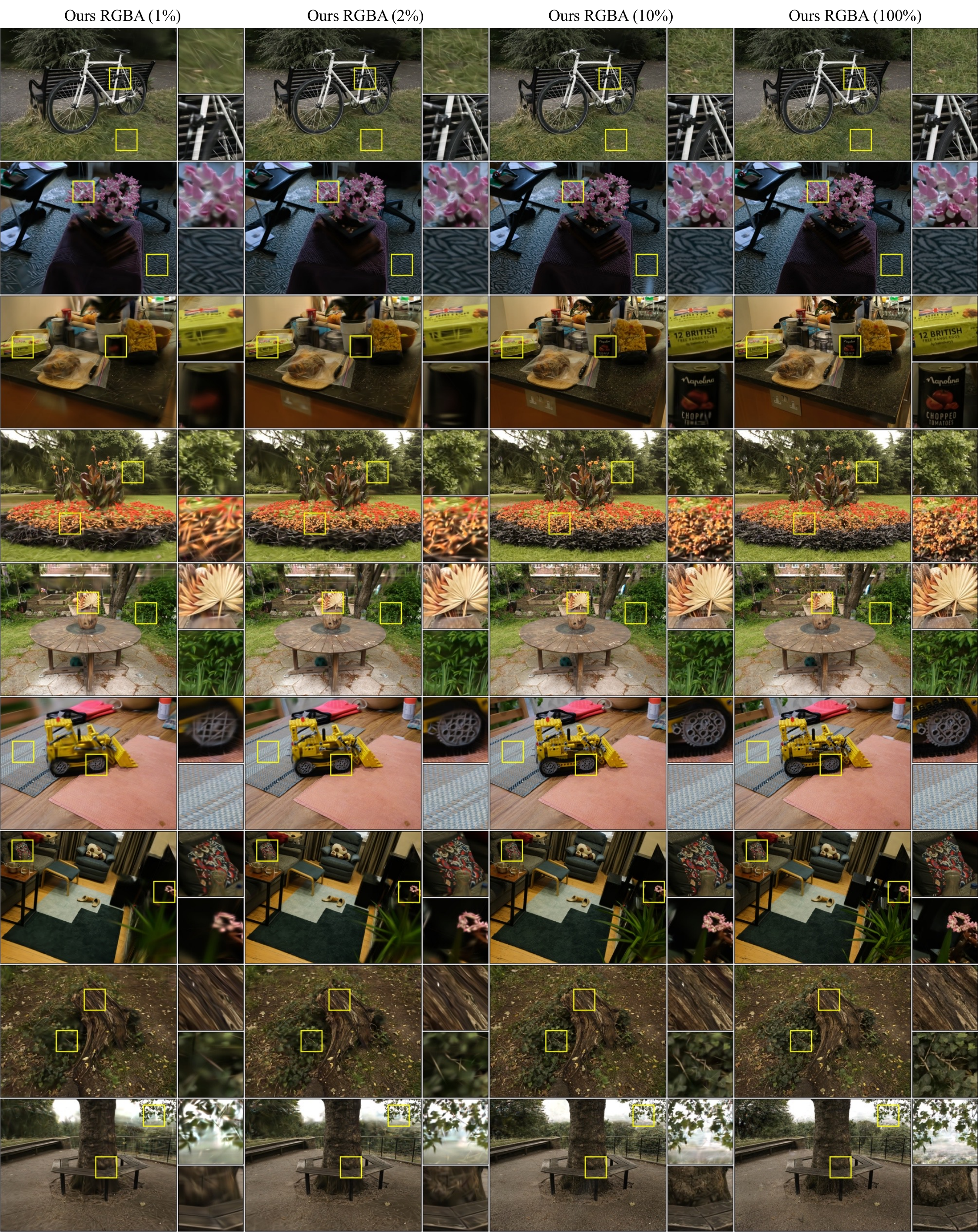}
    \caption{\textbf{Qualitative NVS results on the Mip-NeRF 360 dataset with vayring numbers of Gaussians and a fixed texture map resolution. } Given a fixed texture map resolution, novel-view synthesis performance improves as the number of Gaussians increases. Detailed appearance and complex geometry can be reconstructed with more and therefore smaller Gaussians. }
    \label{fig:vary_num_gs_mipnerf360}
\end{figure*}

\begin{figure*}
    \centering
    \includegraphics[width=\linewidth]{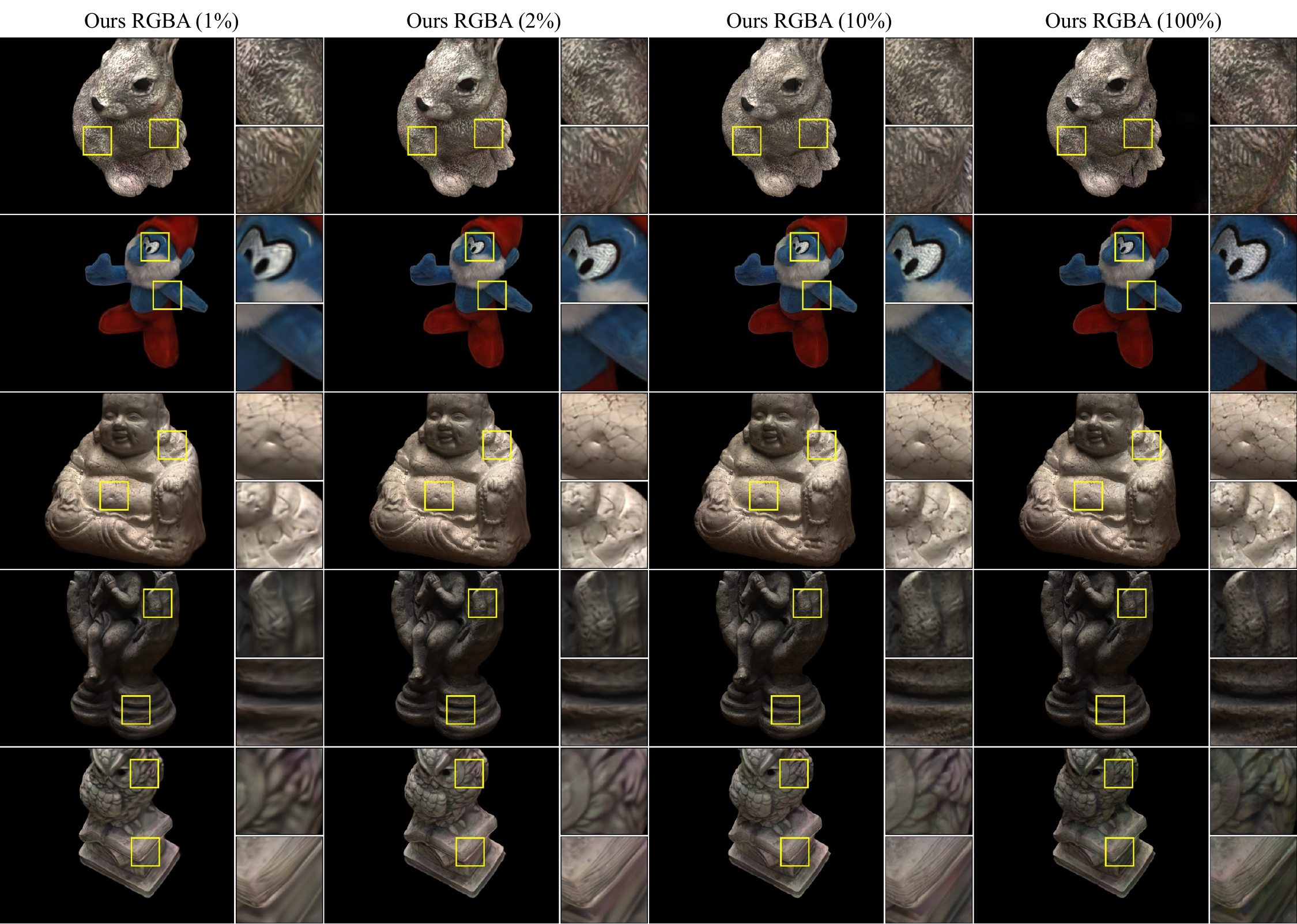}
    \caption{\textbf{Qualitative NVS results on the DTU dataset with vayring numbers of Gaussians and a fixed texture map resolution. } Given a fixed texture map resolution, novel-view synthesis performance improves as the number of Gaussians increases. Detailed appearance and complex geometry can be reconstructed with more and therefore smaller Gaussians. }
    \label{fig:vary_num_gs_dtu}
\end{figure*}

\begin{figure*}
    \centering
    \includegraphics[width=\linewidth]{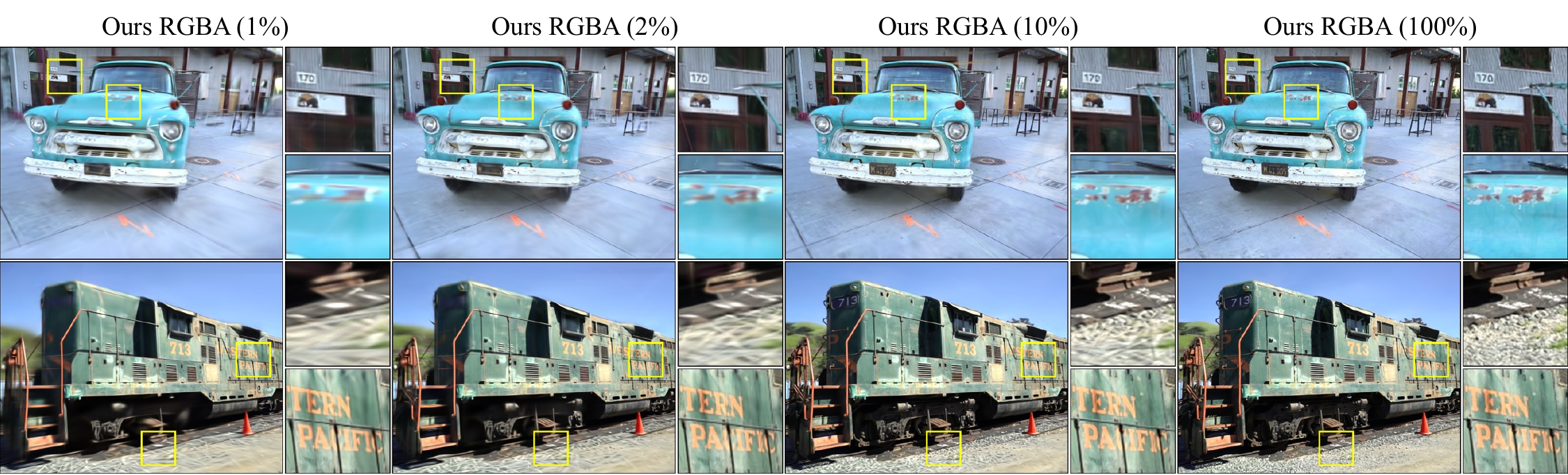}
    \caption{\textbf{Qualitative NVS results on the Tanks and Temples dataset with vayring numbers of Gaussians and a fixed texture map resolution. } Given a fixed texture map resolution, novel-view synthesis performance improves as the number of Gaussians increases. Detailed appearance and complex geometry can be reconstructed with more and therefore smaller Gaussians. }
    \label{fig:vary_num_gs_tandt}
\end{figure*}

\begin{figure*}
    \centering
    \includegraphics[width=\linewidth]{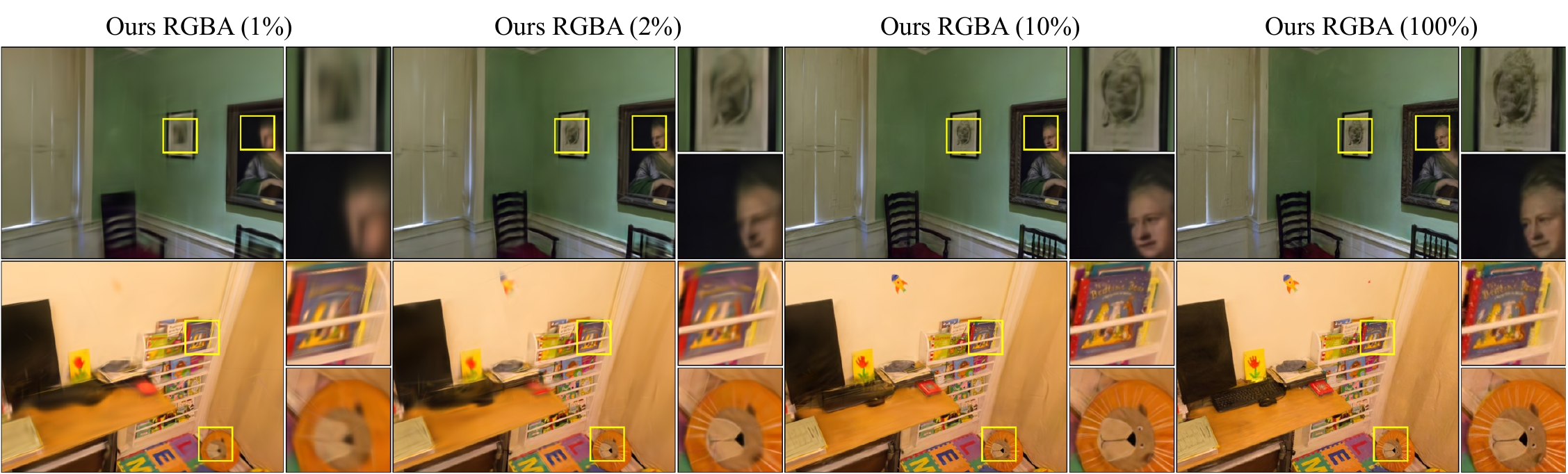}
    \caption{\textbf{Qualitative NVS results on the Deep Blending dataset with vayring numbers of Gaussians and a fixed texture map resolution. } Given a fixed texture map resolution, novel-view synthesis performance improves as the number of Gaussians increases. Detailed appearance and complex geometry can be reconstructed with more and therefore smaller Gaussians. }
    \label{fig:vary_num_gs_db}
\end{figure*}

\begin{figure*}
    \centering
    \includegraphics[width=\linewidth]{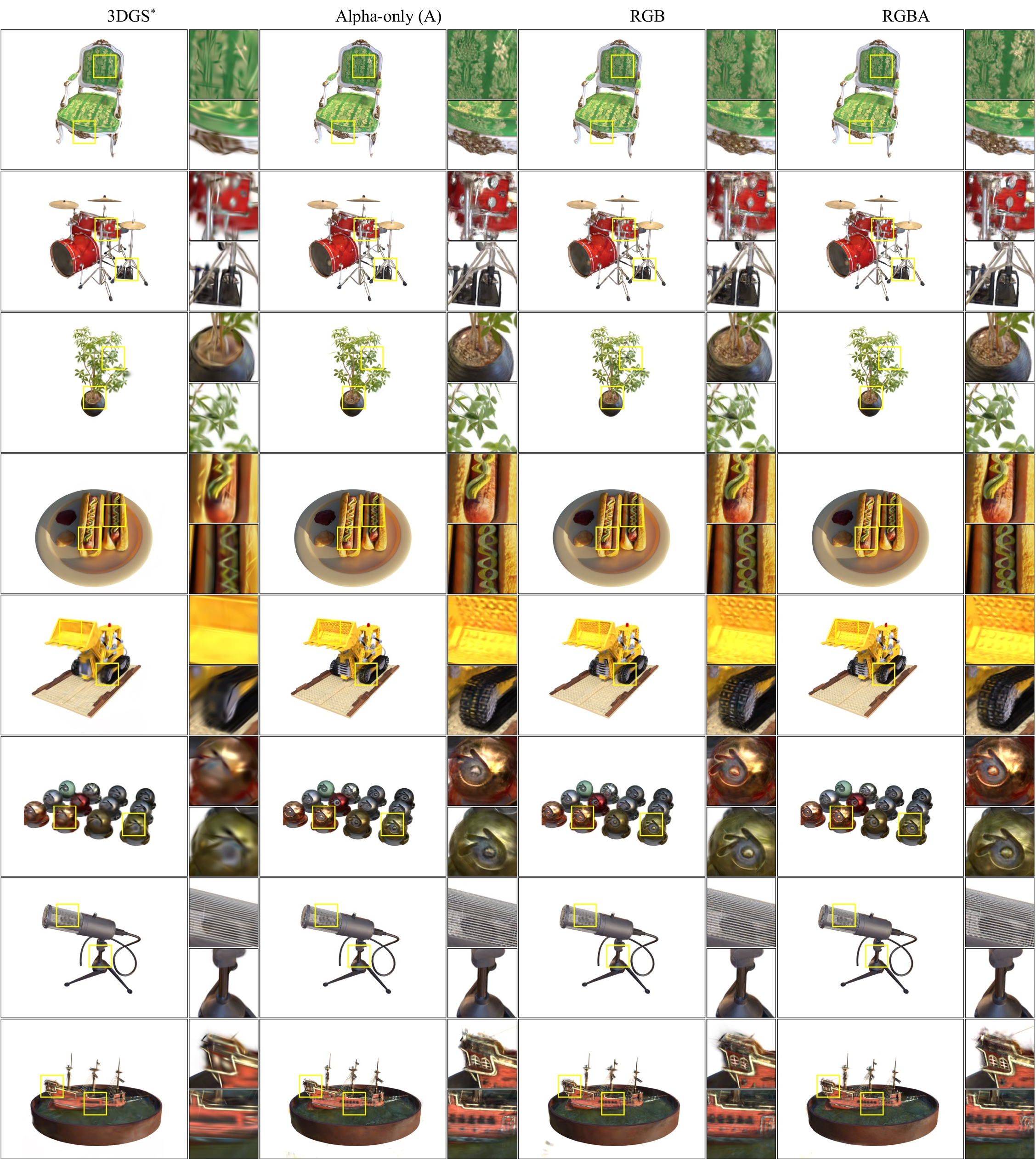}
    \caption{\textbf{Texture map ablation results on the Blender dataset. } We see that models with alpha textures (alpha-only and RGBA models) achieves better NVS quality since \textit{both} appearance and geometry can be reconstructed better due to spatially varying alpha modulation and composition. On the other hand, RGB Textured Gaussians models achieve worse visual quality since each Gaussian can still only represent ellipsoids. }
    \label{fig:tex_ablation_blender}
\end{figure*}

\begin{figure*}
    \centering
    \includegraphics[height=0.9 \textheight]{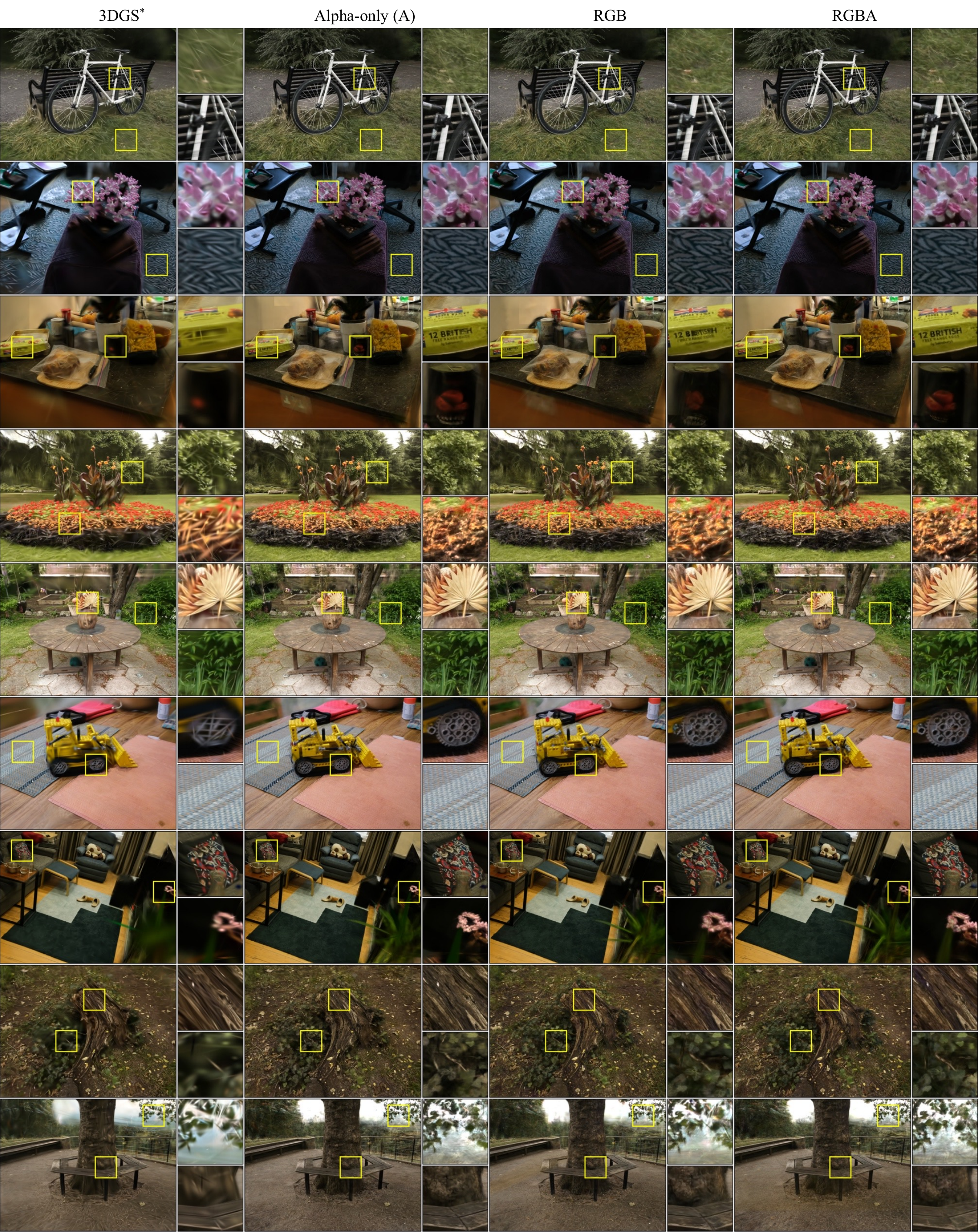}
    \caption{\textbf{Texture map ablation results on the Mip-NeRF 360 dataset. }We see that models with alpha textures (alpha-only and RGBA models) achieves better NVS quality since \textit{both} appearance and geometry can be reconstructed better due to spatially varying alpha modulation and composition. On the other hand, RGB Textured Gaussians models achieve worse visual quality since each Gaussian can still only represent ellipsoids. }
    \label{fig:tex_ablation_mipnerf360}
\end{figure*}

\begin{figure*}
    \centering
    \includegraphics[width=\linewidth]{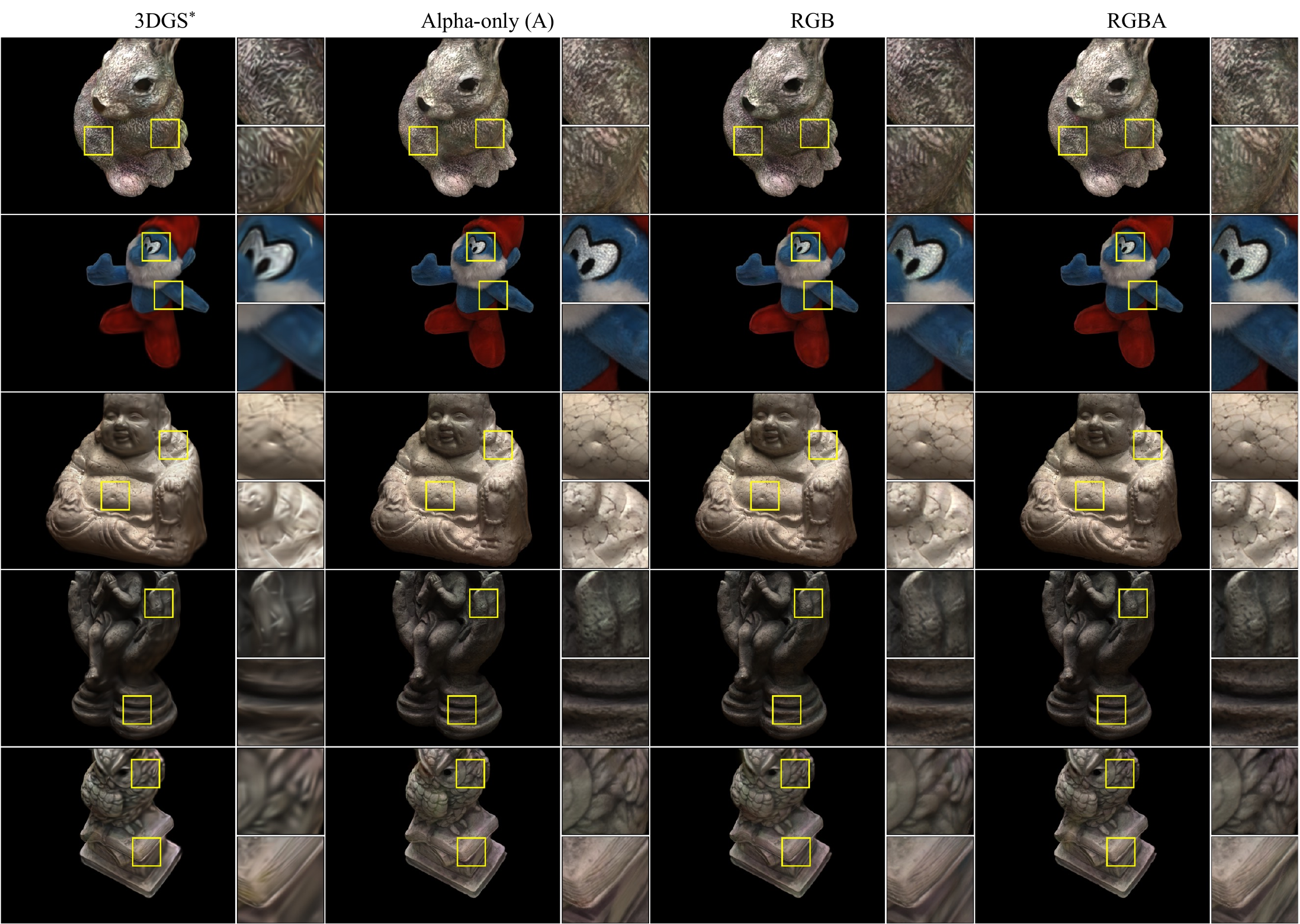}
    \caption{\textbf{Texture map ablation results on the DTU dataset. }We see that models with alpha textures (alpha-only and RGBA models) achieves better NVS quality since \textit{both} appearance and geometry can be reconstructed better due to spatially varying alpha modulation and composition. On the other hand, RGB Textured Gaussians models achieve worse visual quality since each Gaussian can still only represent ellipsoids. }
    \label{fig:tex_ablation_dtu}
\end{figure*}

\begin{figure*}
    \centering
    \includegraphics[width=\linewidth]{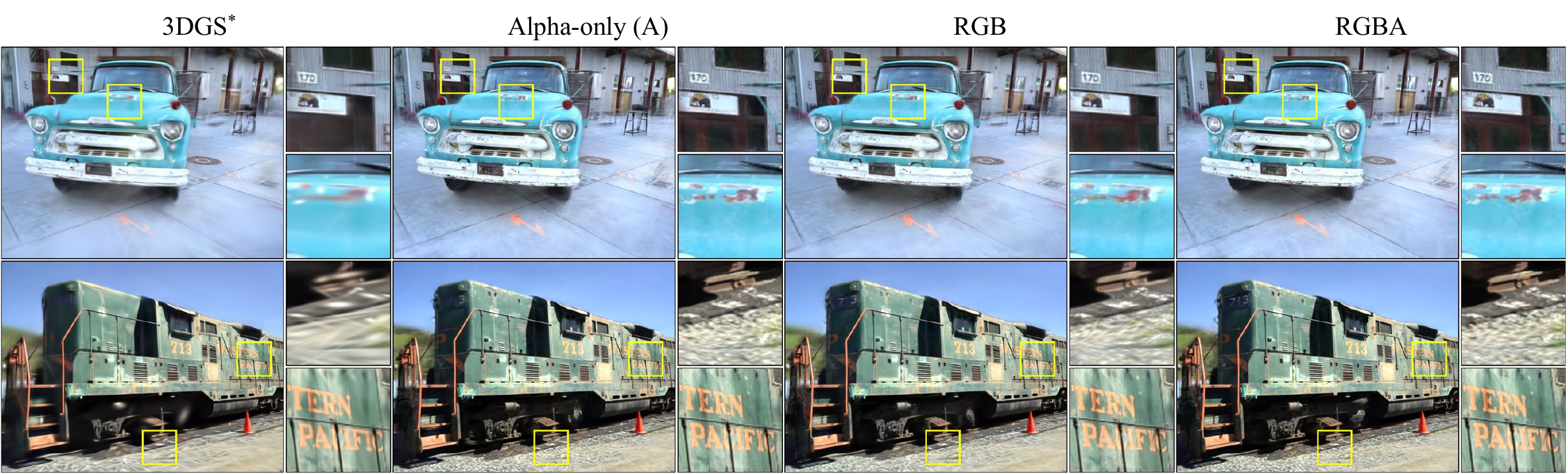}
    \caption{\textbf{Texture map ablation results on the Tanks and Temples dataset. }We see that models with alpha textures (alpha-only and RGBA models) achieves better NVS quality since \textit{both} appearance and geometry can be reconstructed better due to spatially varying alpha modulation and composition. On the other hand, RGB Textured Gaussians models achieve worse visual quality since each Gaussian can still only represent ellipsoids. }
    \label{fig:tex_ablation_tandt}
\end{figure*}

\begin{figure*}
    \centering
    \includegraphics[width=\linewidth]{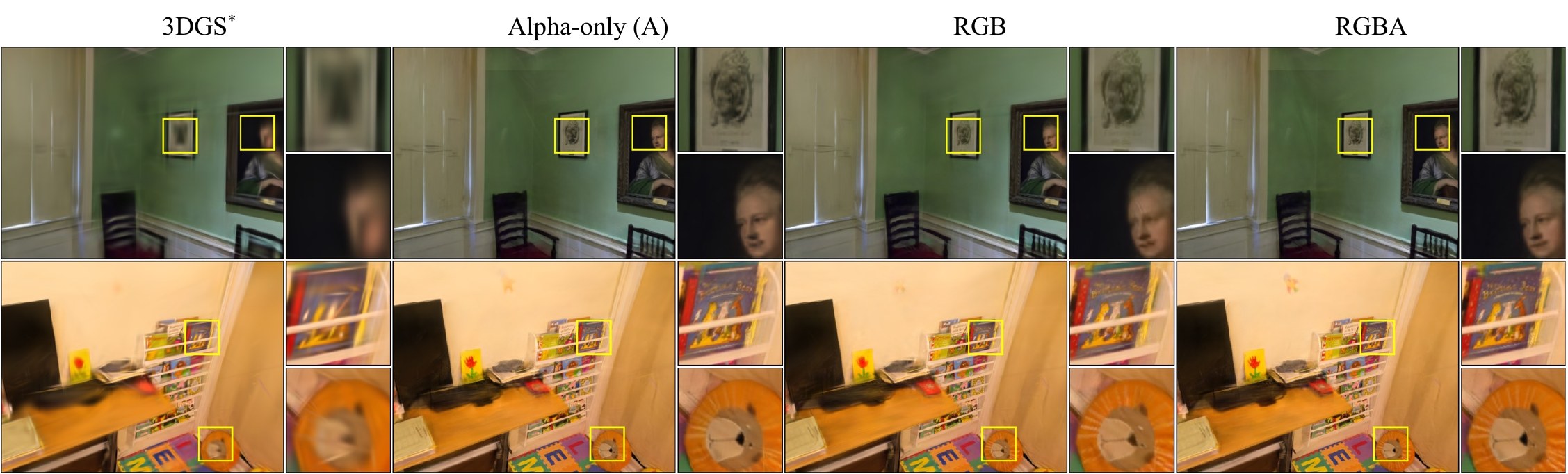}
    \caption{\textbf{Texture map ablation results on the Deep Blending dataset. }We see that models with alpha textures (alpha-only and RGBA models) achieves better NVS quality since \textit{both} appearance and geometry can be reconstructed better due to spatially varying alpha modulation and composition. On the other hand, RGB Textured Gaussians models achieve worse visual quality since each Gaussian can still only represent ellipsoids. }
    \label{fig:tex_ablation_db}
\end{figure*}

\begin{figure*}
    \centering
    \includegraphics[height=0.9 \textheight]{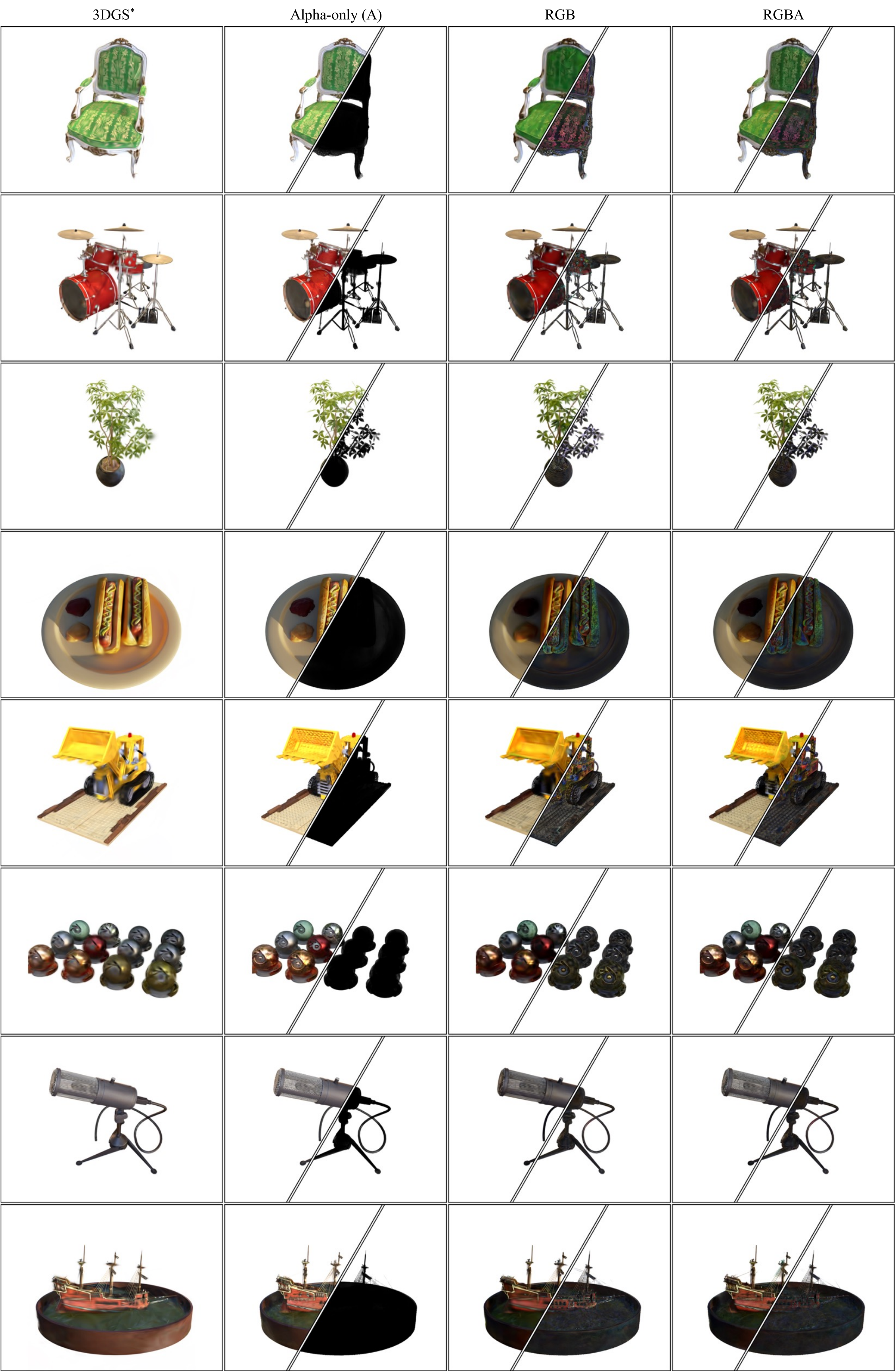}
    \caption{\textbf{Color component decomposition visualization of the Blender dataset. } With alpha textures, the alpha-modulated and composited base color component can already recosntruct high-frequency textures, as shown in results of the alpha-only and RGBA Texuted Gaussians models. For models with only RGB textures, the base color component reconstructs lower frequency colors while the texture map color component reconstructs high frequency appearance. }
    \label{fig:color_contribution_blender}
\end{figure*}

\begin{figure*}
    \centering
    \includegraphics[height=0.9 \textheight]{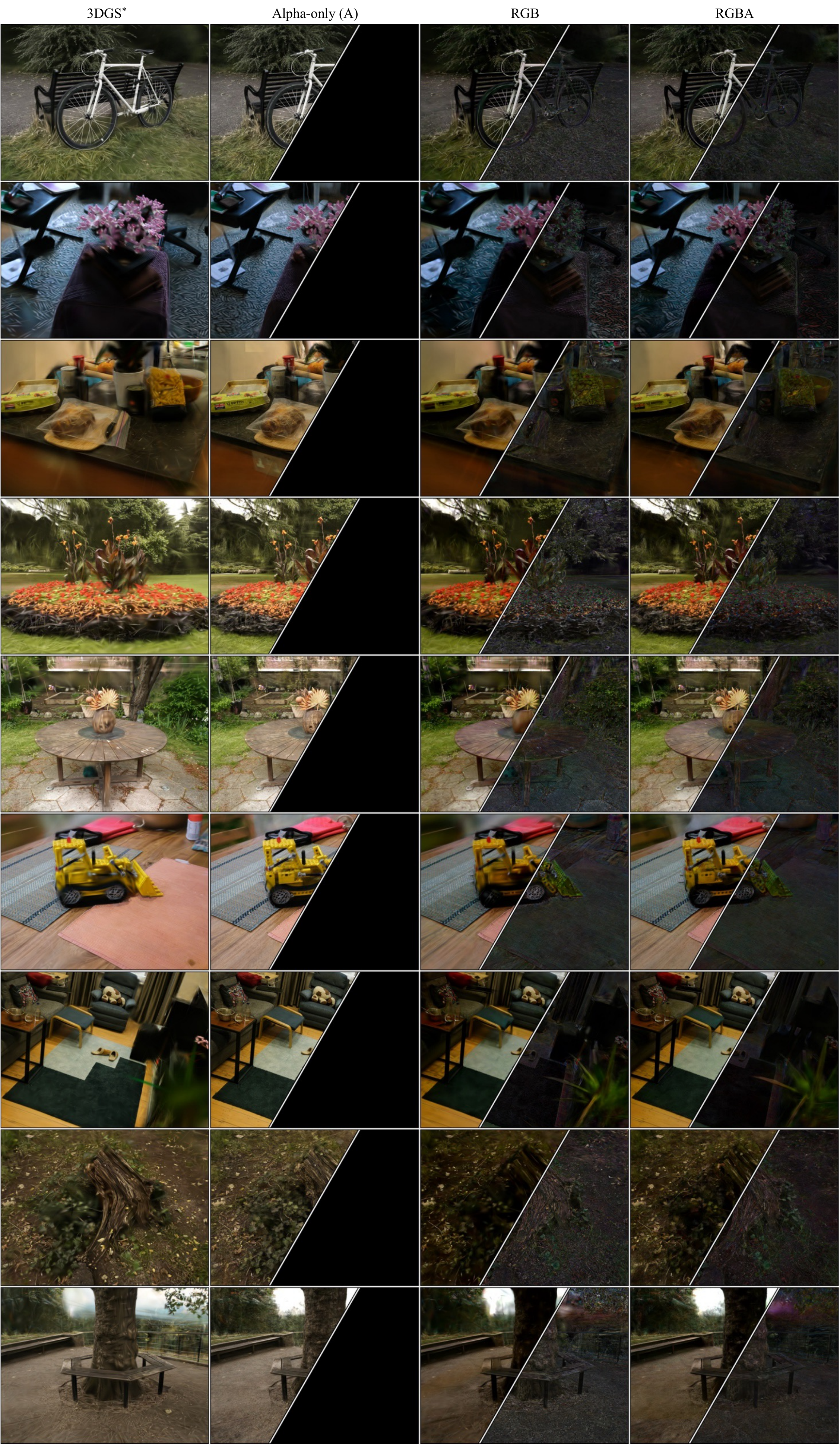}
    \caption{\textbf{Color component decomposition visualization of the Mip-NeRF 360 dataset. } With alpha textures, the alpha-modulated and composited base color component can already recosntruct high-frequency textures, as shown in results of the alpha-only and RGBA Texuted Gaussians models. For models with only RGB textures, the base color component reconstructs lower frequency colors while the texture map color component reconstructs high frequency appearance. }
    \label{fig:color_contribution_mipnerf360}
\end{figure*}

\begin{figure*}
    \centering
    \includegraphics[width=\linewidth]{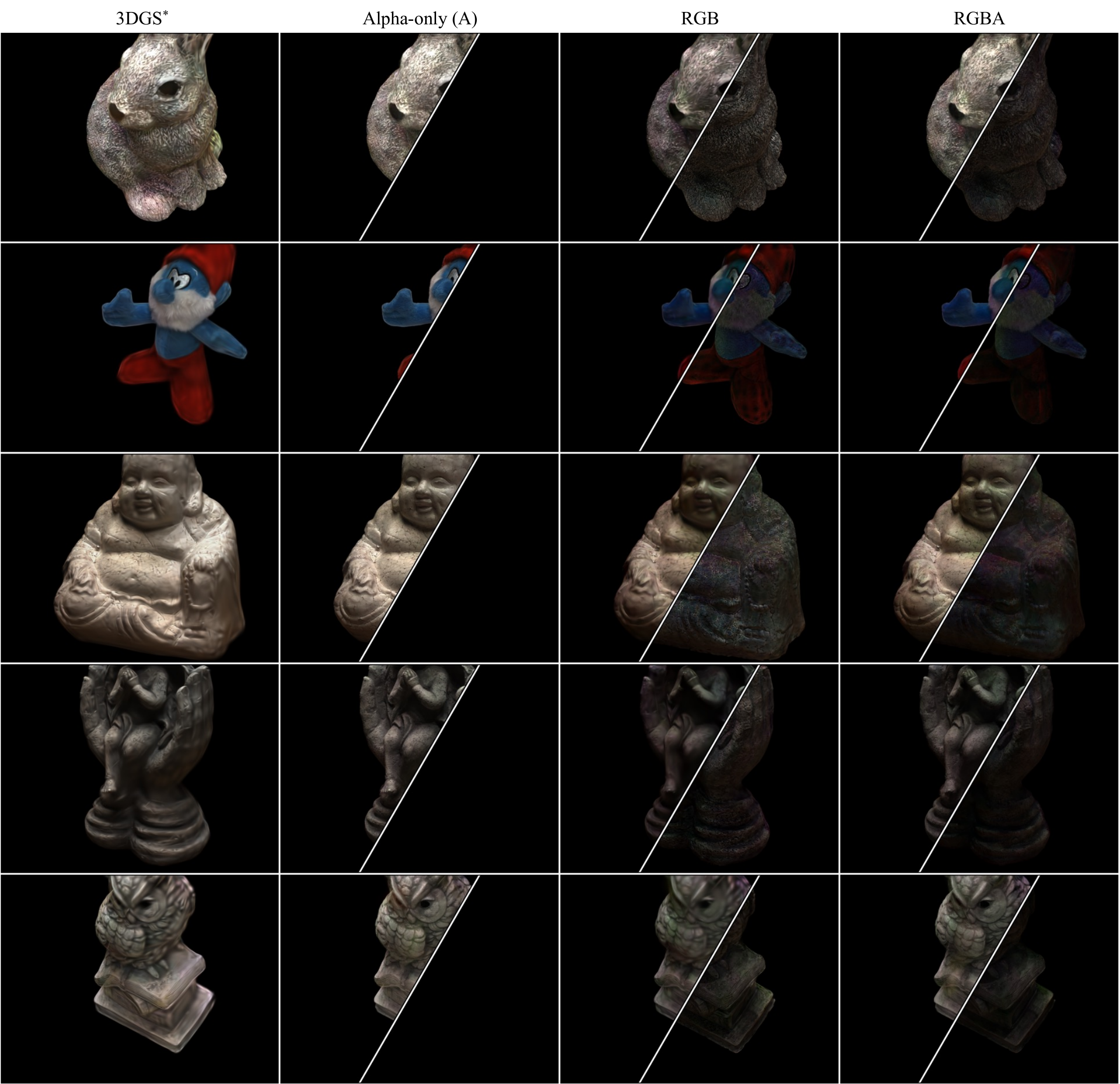}
    \caption{\textbf{Color component decomposition visualization of the DTU dataset. } With alpha textures, the alpha-modulated and composited base color component can already recosntruct high-frequency textures, as shown in results of the alpha-only and RGBA Texuted Gaussians models. For models with only RGB textures, the base color component reconstructs lower frequency colors while the texture map color component reconstructs high frequency appearance. }
    \label{fig:color_contribution_dtu}
\end{figure*}

\begin{figure*}
    \centering
    \includegraphics[width=\linewidth]{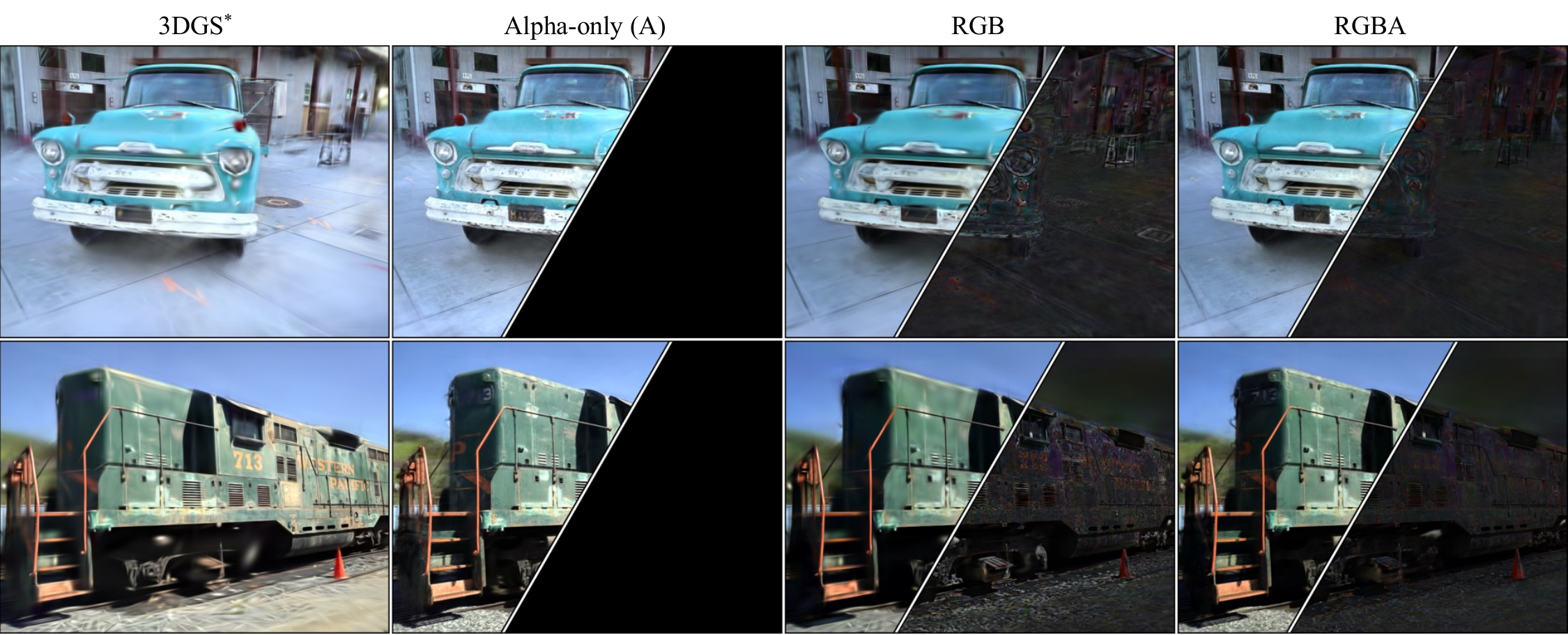}
    \caption{\textbf{Color component decomposition visualization of the Tanks and Temples dataset. } With alpha textures, the alpha-modulated and composited base color component can already recosntruct high-frequency textures, as shown in results of the alpha-only and RGBA Texuted Gaussians models. For models with only RGB textures, the base color component reconstructs lower frequency colors while the texture map color component reconstructs high frequency appearance. }
    \label{fig:color_contribution_tandt}
\end{figure*}

\begin{figure*}
    \centering
    \includegraphics[width=\linewidth]{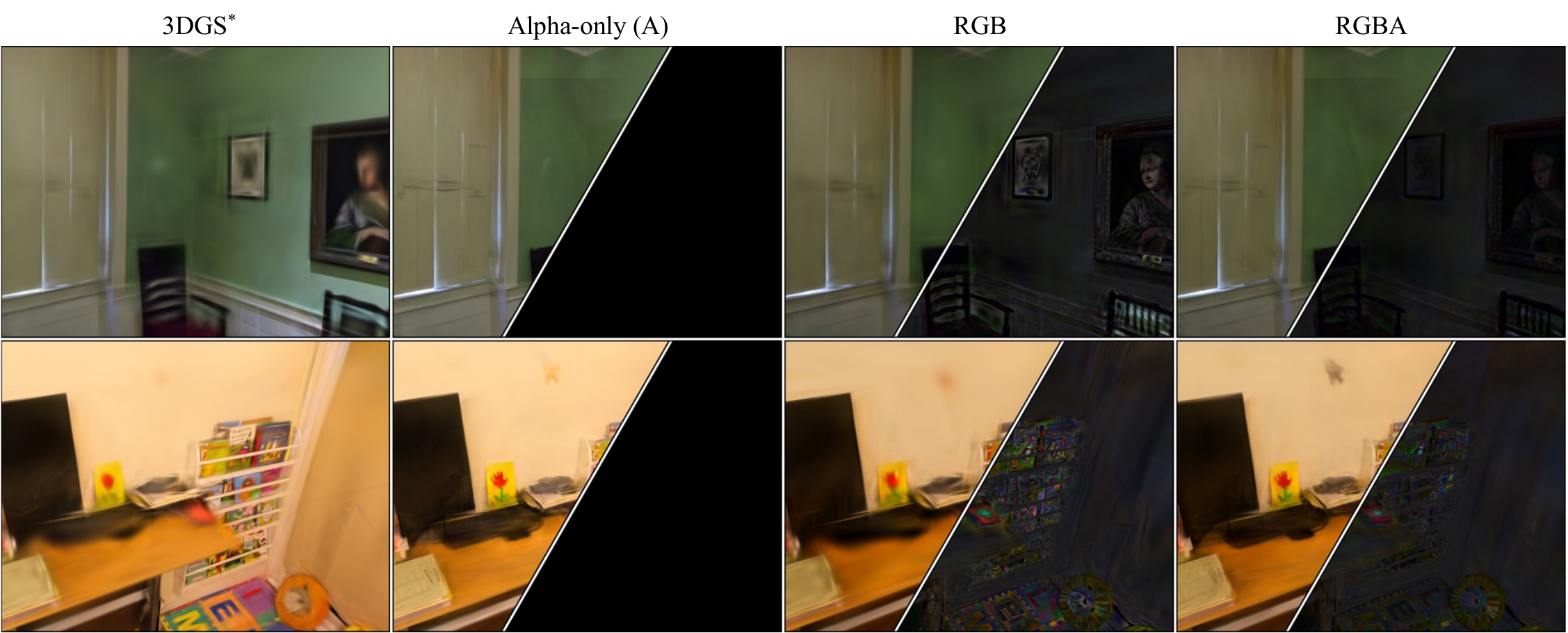}
    \caption{\textbf{Color component decomposition visualization of the Deep Blending dataset. } With alpha textures, the alpha-modulated and composited base color component can already recosntruct high-frequency textures, as shown in results of the alpha-only and RGBA Texuted Gaussians models. For models with only RGB textures, the base color component reconstructs lower frequency colors while the texture map color component reconstructs high frequency appearance. }
    \label{fig:color_contribution_db}
\end{figure*}

We also used various designs to reduce artifacts especially when rendering flat Gaussians from grazing viewpoints. As shown in Figure \ref{fig:intersection_ablation}, intersecting Gaussians only within the $\pm3\sigma$ range and using SH base colors greatly reduce rendering artifacts. 

{\renewcommand{\arraystretch}{1.2}
\begin{table*}[ht!]
  \centering
  \footnotesize
  \begin{tabular*}{\textwidth}{@{\extracolsep{\fill}}lccccc@{}}
    \toprule
    \makecell{Method} & \makecell{Blender \cite{mildenhall2020nerf}} & \makecell{Mip-NeRF 360 \cite{mildenhall2020nerf}}  & \makecell{DTU \cite{jensen2014DTU}} & \makecell{Tanks and Temples \cite{Knapitsch2017tanks}} & \makecell{Deep Blending \cite{DeepBlending2018}} \\
    \midrule
    3DGS* ($1\%$) & 26.89 / 0.9160 / 0.1165 & 22.37 / 0.6293 / 0.4774 & 30.88 / 0.9320 / 0.1581 & 19.90 / 0.6736 / 0.4406 & 23.97 / 0.8167 / 0.4337
 \\
    Ours ($1\%$) & 28.02 / 0.9340 / 0.0847 & 23.75 / 0.7066 / 0.3367 & 32.41 / 0.9626 / 0.0696 & 21.08 / 0.7384 / 0.3114 & 24.88 / 0.8454 / 0.3550

 \\
\hdashline
    3DGS* ($2\%$) & 28.44 / 0.9320 / 0.0973 & 23.43 / 0.6816 / 0.4216 & 31.90 / 0.9478 / 0.1317 & 21.05 / 0.7199 / 0.3881 & 25.49 / 0.8459 / 0.3894
\\
    Ours ($2\%$) & 29.68 / 0.9465 / 0.0716 & 24.79 / 0.7454 / 0.2945 & 32.77 / 0.9652 / 0.0621 & 22.14 / 0.7757 / 0.2726 & 26.29 / 0.8681 / 0.3186
\\
\hdashline
    3DGS* ($5\%$) & 30.09 / 0.9491 / 0.0742 & 24.88 / 0.7450 / 0.3393 & 32.63 / 0.9593 / 0.0987 & 22.25 / 0.7720 / 0.3189 & 26.55 / 0.8642 / 0.3469
 \\
    Ours ($5\%$) & 30.96 / 0.9564 / 0.0571 & 25.87 / 0.7841 / 0.2492 & 32.96 / 0.9662 / 0.0584 & 23.05 / 0.8085 / 0.2339 & 27.00 / 0.8735 / 0.3039
\\
\hdashline
    3DGS* ($10\%$) & 31.47 / 0.9590 / 0.0594 & 25.77 / 0.7796 / 0.2860 & 32.71 / 0.9627 / 0.0811 & 22.82 / 0.8020 / 0.2728 & 27.64 / 0.8853 / 0.3101
 \\
    Ours ($10\%$) & 32.12 / 0.9630 / 0.0488 & 26.45 / 0.8012 / 0.2298 & 32.72 / 0.9659 / 0.0583 & 23.42 / 0.8258 / 0.2123 & 27.98 / 0.8899 / 0.2803
\\
\hdashline
    3DGS* ($20\%$) & 32.21 / 0.9637 / 0.0506 & 26.45 / 0.8057 / 0.2408 & 32.93 / 0.9651 / 0.0699 & 23.58 / 0.8305 / 0.2298 & 27.99 / 0.8908 / 0.2927
\\
    Ours ($20\%$) & 32.59 / 0.9656 / 0.0453 & 26.83 / 0.8139 / 0.2138 & 32.85 / 0.9666 / 0.0584 & 23.99 / 0.8428 / 0.1940 & 28.16 / 0.8906 / 0.2784
\\
\hdashline
    3DGS* ($50\%$) & 32.77 / 0.9664 / 0.0452 & 27.02 / 0.8250 / 0.2019 & 33.23 / 0.9683 / 0.0603 & 23.96 / 0.8479 / 0.1922 & 28.11 / 0.8939 / 0.2762
 \\
    Ours ($50\%$) & 32.98 / 0.9673 / 0.0434 & 27.13 / 0.8245 / 0.1973 & 33.19 / 0.9686 / 0.0594 & 24.18 / 0.8516 / 0.1801 & 28.27 / 0.8915 / 0.2738
\\
\hdashline
    3DGS* ($100\%$) & 33.09 / 0.9671 / 0.0440 & 27.28 / 0.8318 / 0.1871 & 33.54 / 0.9697 / 0.0551 & 24.18 / 0.8541 / 0.1754 & 28.04 / 0.8940 / 0.2707
 \\
    Ours ($100\%$) & 33.24 / 0.9674 / 0.0428 & 27.35 / 0.8274 / 0.1858 & 33.61 / 0.9699 / 0.0556 & 24.26 / 0.8542 / 0.1684 & 28.33 / 0.8908 / 0.2699
\\
    \bottomrule
  \end{tabular*}
  \caption{\textbf{Quantitaive results on all datasets using varying numbers of Gaussians with a fixed amount of texels. } We report the quantitative NVS performance of our RGBA Textured Gaussians model and 3DGS$^*$ with the same number of Gaussians on all five test datasets in terms of  PSNR ($\uparrow$), SSIM ($\uparrow$), and LPIPS ($\downarrow$) metrics. Our Textured Gaussians model consistently outperforms 3DGS$^*$ when using different numbers of Gaussians. }
  \label{tab:supp_same_num_gs}
\end{table*}
}

\clearpage
\newpage

\clearpage
\newpage

\clearpage
\newpage
{\renewcommand{\arraystretch}{1.2}
\begin{table*}[ht!]
  \centering
  \footnotesize
  \begin{tabular*}{\textwidth}{@{\extracolsep{\fill}}lccccc@{}}
    \toprule
    \makecell{Method} & \makecell{Blender \cite{mildenhall2020nerf}} & \makecell{Mip-NeRF 360 \cite{mildenhall2020nerf}}  & \makecell{DTU \cite{jensen2014DTU}} & \makecell{Tanks and Temples \cite{Knapitsch2017tanks}} & \makecell{Deep Blending \cite{DeepBlending2018}} \\
    \midrule
    3DGS* & 33.09 / 0.9671 / 0.0440 & 27.28 / 0.8318 / 0.1871 & 33.54 / 0.9697 / 0.0551 & 24.18 / 0.8541 / 0.1754 & 28.04 / 0.8940 / 0.2707
 \\
    Alpha-only  & 33.22 / 0.9672 / 0.0433 & 27.32 / 0.8259 / 0.1874 & 33.51 / 0.9692 / 0.0554 & 28.30 / 0.8888 / 0.2720 & 24.27 / 0.8532 / 0.1695
 \\ 
    RGB  & 33.20 / 0.9673 / 0.0431 & 27.30 / 0.8268 / 0.1873 & 33.58 / 0.9697 / 0.0561 & 28.21 / 0.8895 / 0.2722 & 24.24 / 0.8536 / 0.1695
 \\
    RGBA  & 33.24 / 0.9674 / 0.0429 & 27.35 / 0.8274 / 0.1859 & 33.60 / 0.9699 / 0.0559 & 28.34 / 0.8910 / 0.2699 & 24.26 / 0.8542 / 0.1685
 \\
    \hline 
    3DGS* (1\%)& 26.89 / 0.9160 / 0.1165 & 22.37 / 0.6293 / 0.4774 & 30.88 / 0.9320 / 0.1581 & 19.90 / 0.6736 / 0.4406 & 23.97 / 0.8167 / 0.4337
 \\
    Alpha-only (1\%)& 27.64 / 0.9304 / 0.0905 & 23.69 / 0.7012 / 0.3494 & 32.31 / 0.9604 / 0.0759 & 24.68 / 0.8411 / 0.3620 & 20.93 / 0.7335 / 0.3226
 \\ 
    RGB (1\%)& 27.60 / 0.9279 / 0.0923 & 23.50 / 0.6971 / 0.3558 & 32.52 / 0.9612 / 0.0765 & 24.64 / 0.8403 / 0.3682 & 20.73 / 0.7236 / 0.3352
 \\
    RGBA (1\%)& 28.11 / 0.9343 / 0.0849 & 23.73 / 0.7064 / 0.3365 & 32.43 / 0.9627 / 0.0694 & 24.83 / 0.8454 / 0.3552 & 21.08 / 0.7395 / 0.3107
 \\
    \bottomrule
  \end{tabular*}
  \caption{\textbf{Quantitative results of ablation study on texture map variants. } We ablate different variants of our model that use alpha, RGB, and RGBA texture maps with the default optimized number and 1\% of the default optimized number of Gaussians. We report the NVS performance of different models in terms of PSNR ($\uparrow$), SSIM ($\uparrow$), and LPIPS ($\downarrow$) metrics. Models with RGBA textures achieve the best results due to the maximum expressivity of individual Gaussians. Interestingly, models with alpha-only textures achieve better results than RGB textures models while using only one-third of the model size. }
  \label{tab:supp_tex_ablation}
\end{table*}
}

